%% file: psf-2022550-english.tex
\documentclass[preprints,article,accept,pdftex,moreauthors]{Definitions/mdpi} 
\firstpage{1} 
\pubvolume{5}
\issuenum{1}
\articlenumber{12}
\pubyear{2022}
\copyrightyear{2022}
\externaleditor{Academic Editors: Frédéric
Barbaresco, Ali Mohammad-Djafari,
Frank Nielsen and Martino \newline 
Trassinelli}
\datereceived{} 
\dateaccepted{} 
\datepublished{
} 
\hreflink{
} 
\doinum{
}
\pdfoutput=1

\usepackage{bm} 			
\usepackage{tikz}			
\usepackage{pgfplots}		
\usepgfplotslibrary{ternary}
\usepackage{amsmath}		
\usepackage[linesnumbered,ruled,vlined]{algorithm2e}
\usepackage{filecontents}
\usepackage{fontawesome}
\usepackage{bbold}
\usepackage{subfig}
\usepackage{cleveref}
\usepackage{multirow}
\usepackage[utf8]{inputenc} 
\usepackage[T1]{fontenc} 
\usepackage{url}  
\usepackage{booktabs} 
\usepackage{amsfonts} 
\usepackage{nicefrac} 
\usepackage{microtype} 
\usepackage{algpseudocode}
\usepackage{longtable}
\usepackage{float}
\usepackage{mathtools}

\usetikzlibrary{shapes.geometric}
\usetikzlibrary{arrows,positioning}
\usetikzlibrary{patterns}
\usetikzlibrary{matrix,decorations.pathreplacing}
\usetikzlibrary{positioning}
\usetikzlibrary{arrows,decorations.markings}
\usetikzlibrary{decorations.pathreplacing}
\usetikzlibrary{backgrounds}
\usetikzlibrary{fit}
\usetikzlibrary{shapes.geometric}
\usetikzlibrary{arrows,positioning}
\usetikzlibrary{patterns}
\usetikzlibrary{matrix,decorations.pathreplacing}
\usetikzlibrary{positioning}
\usetikzlibrary{arrows,decorations.markings}
\usetikzlibrary{decorations.pathreplacing}
\usetikzlibrary{pgfplots.groupplots}

\definecolor{Red}{rgb}{1.0, 0.03, 0.0}
\definecolor{Blue}{rgb}{0.0, 0.5, 1.0}
\definecolor{Green}{rgb}{0.0, 0.42, 0.24}
\definecolor{Brown}{rgb}{0.63, 0.47, 0.35}
\definecolor{Orange}{rgb}{1.0, 0.65, 0.0}
\definecolor{Gray}{rgb}{0.66, 0.66, 0.66}
\definecolor{Purple}{rgb}{0.6, 0.2, 0.8}
\definecolor{Cyan}{rgb}{0.0, 0.81, 0.82}
\definecolor{VioletRed}{rgb}{1.0, 0.08, 0.58}
\definecolor{Black}{rgb}{0.0, 0.0, 0.0}

\newcommand*{\equal}{=}
\newcommand*\circled[1]{\tikz[baseline=(char.base)]{
		\node[shape=circle,draw,inner sep=1.2pt] (char) {#1};}}

\Title{Classification and Uncertainty Quantification of Corrupted Data Using Supervised Autoencoders 
}

\TitleCitation{Classification and Uncertainty Quantification of Corrupted Data Using Supervised Autoencoders}

\Author{Philipp Joppich $^{1,2,}$, Sebastian Dorn $^{2,}$, Oliver De Candido $^1$, Jakob Knollmüller $^{3,4}$\linebreak and Wolfgang Utschick $^1$} 

\AuthorNames{Philipp Joppich, Sebastian Dorn, Oliver De Candido, Jakob Knollmüller and Wolfgang Utschick}

\AuthorCitation{Joppich, P.; Dorn, S.;\newline
De Candido, O.; Knollmüller, J.;
Utschick, W.}

\address{%
$^1$ \quad School of Computation, Information and Technology, Technical University of Munich, Arcisstr. 21,\newline
80333 Munich, Germany

$^2$ \quad AUDI AG, Auto-Union-Str. 1, 85057 Ingolstadt, Germany

$^3$ \quad School of Natural Sciences, Technical University of Munich, James-Franck-Str. 1, 85748 Garching, Germany

$^4$ \quad Excellence Cluster ORIGINS, Boltzmannstr. 2, 85748 Garching, Germany
}

\corres{Correspondence: philipp.joppich@tum.de (P.J.); sebastian.dorn@audi.de (S.D.)}

\abstract{Parametric and non-parametric classifiers often have to deal with real-world data, where corruptions such as noise, occlusions, and blur are unavoidable. We present a probabilistic approach to classify strongly corrupted data and quantify uncertainty, even though the corrupted data do not have to be included to the training data. A supervised autoencoder is the underlying architecture. We used the decoding part as a generative model for realistic data and extended it by convolutions, masking, and additive Gaussian noise to describe imperfections. This constitutes a statistical inference task in terms of the optimal latent space activations of the underlying uncorrupted datum. We solved this problem approximately with Metric Gaussian Variational Inference (MGVI). The supervision of the autoencoder's latent space allowed us to classify corrupted data directly under uncertainty with the statistically inferred latent space activations. We show that the derived model uncertainty can be used as a statistical ``lie detector'' of the classification. Independent of that, the generative model can optimally restore the corrupted datum by decoding the inferred latent space activations.}

\begin{document}
\section{Introduction and Motivation}

\begin{figure}[H]
\centering
\input{images/figure1}

	\caption{From left to right: Ground truth image $\bm{x}$ in the data space, corrupted image $\bm{d}$ in the data space (random masking $\bm{m}$, Gaussian blur $\bm{C}$, additive white Gaussian noise $\bm{n}$), posterior mean $\overline{\mathcal{H}}$ in the latent space with reconstruction uncertainty $\bm{\delta}_r$, model uncertainty $\bm{\delta}_m$, and the restored image $g(\overline{\mathcal{H}})$ (decoded posterior mean) in the data space. We included the encoding of the uncorrupted data $f(\bm{x})$ (illustrated by the shaded white bars in the third column). Top row: data sample from the MNIST-dataset (ground truth label: $4$). Bottom row: data sample from the Fashion-MNIST-dataset (ground truth label: 2 (pullover)). We can classify $\bm{d}$ using the posterior mean $\overline{\mathcal{H}}$ as the autoencoder's latent space is supervised (note the highlighted max. activation responsible for classification). We are able to classify and quantify model uncertainty $\bm{\delta}_m$ with the Mahalanobis distance in the latent space (note the highlighted min. activation responsible for classification). Strong overlapping for the Fashion-MNIST-example of the $1\cdot\sigma$ error bars of $\bm{\delta}_r$ across different classes indicates that no reliable and confident classification is possible due to heavy corruption.}
	\label{fig:mnist_reconstruction}
\end{figure}
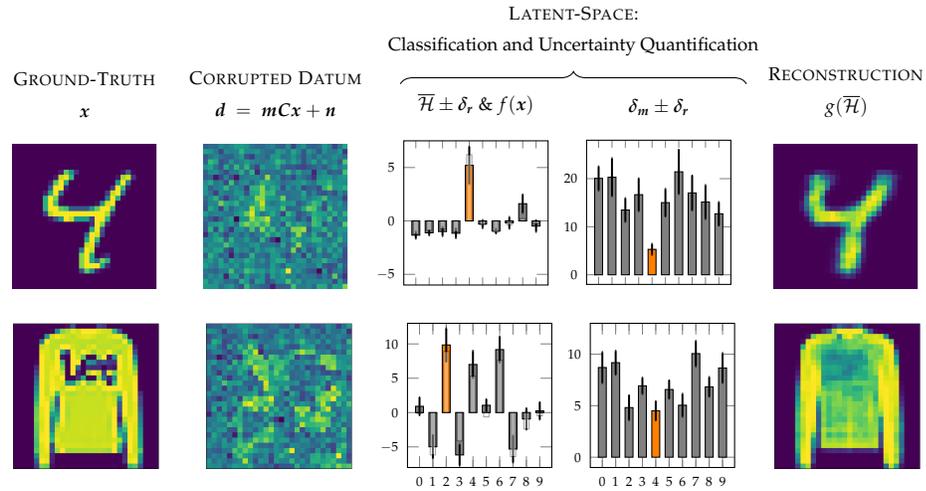

Many real-world applications of data-driven classifiers, e.g., neural networks, involve corruptions that pose significant challenges to the pretrained classifiers. Often, the corruption must previously be included, and, thus, already be known during training. For instance, noise (e.g., due to sensor imperfections) and convolutions (e.g., due to lens flares or unfocused images) are inevitable in image processing systems and may occur spontaneously and irregularly. The same holds for masking, which may occur when a foreign object occludes the actual object of interest (e.g., water droplets, dirt, or scratches on the camera lens).
Hence, we aimed to answer the following question in this paper: How can we classify corrupted data with a parametric classifier without imposing any constraints on the training data? As classifying corrupted data naturally demands a measure of uncertainty for validation, we included both model uncertainty $\bm{\delta}_m$ and reconstruction uncertainty $\bm{\delta}_r$ in the classification. We refer to $\bm{\delta}_m$ as the model's confidence of the classification itself. In contrast, we refer to $\bm{\delta}_r$ as the confidence of the process of reconstructing the latent space activations given some corrupted datum. An overview of the proposed method is illustrated in \Cref{fig:mnist_reconstruction}.
\section{Classification and Uncertainty Quantification of Corrupted Data}
\subsection{Methodology Overview and Related Work}
\label{sec:concept_vis}
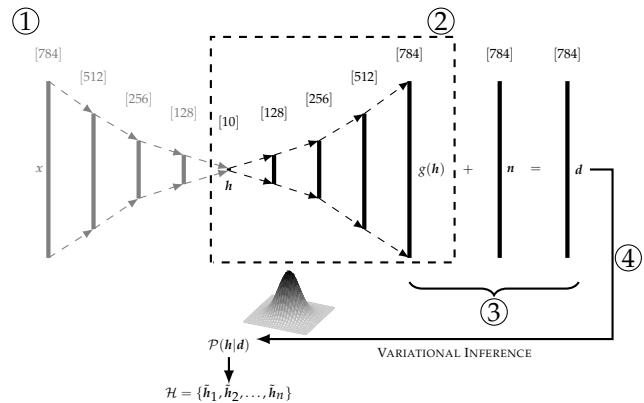
\begin{figure}[H]
\centering
\input{images/figure2}

	\caption{Concept visualization: steps involved in our methodology as described in \Cref{sec:concept_vis}.}
	\label{fig:concept_vis}
\end{figure}

To address the challenge of classification and uncertainty quantification of corrupted data, we propose the following core approach, illustrated in \Cref{fig:concept_vis}.

\circled{1} In the first step, we trained a supervised autoencoder \cite{lei-et-al-2018-sae} that is: (a) capable of classifying the input data with its latent space activations $\bm{h}$ and (b) capable of decoding the (supervised) latent space activations to generate higher-dimensional data, targeting it to be identical to the input. Except for these two constraints ((a) and (b)), we did not impose any further restrictions on the autoencoder. 

\circled{2} In the second step, we decoupled the decoder $g$ from the autoencoder and treated the decoder as a fixed generative function $g$. Neither retraining nor further modifying of $g$ is performed in the following steps.

\circled{3} In the third step, we included $g$ in an \textsc{Additive White Gaussian Noise} (AWGN) channel model $\bm{d}=\bm{m}\bm{C}g(\bm{h})+\bm{n}$. This AWGN channel model additionally involves heavy corruption such as convolution $\bm{C}$ and masking $\bm{m}$. 

\circled{4} In the final step, we approximated the posterior probability distribution $\mathcal{P}(\bm{h}|\bm{d})$ in the latent space and derived the mean and standard deviation, corresponding optimally to some uncorrupted datum $g(\bm{h})$, given the corrupted datum $\bm{d}$. Due to supervision in the latent space, this reconstruction enables a direct classification of $\bm{d}$ including model and reconstruction uncertainty quantification, even though the decoding function was trained on uncorrupted data.
We used a set of samples~$\mathcal{H}$ from the approximate posterior probability distribution to determine the sample mean $\mathrm{mean}(\mathcal{H})=\overline{\mathcal{H}}$, as well as the set's reconstruction uncertainty $\bm{\delta}_r$ with the sample standard deviation $\mathrm{std}(\mathcal{H})$. Samples are statistically inferred by \textsc{Metric Gaussian Variational Inference} (MGVI) \cite{knollmueller-et-al-2020-mgvi}.
In addition to the reconstruction uncertainty $\bm{\delta}_r$, we determined the model uncertainty by calculating the \textsc{Mahalanobis} distance (M-distance) in the latent space representation, slightly different from \cite{lee-et-al-2018-mahalanobis-oodd}. Here, we distinguish between reconstruction uncertainty $\bm{\delta_r}$ and model uncertainty $\bm{\delta_m}$ to evaluate the confidence of the process of inferring $\bm{h}$ and to evaluate the confidence of the classification given by the supervised latent space, respectively. 
Similar to our approach, references \cite{boehm-et-al-2019-uncertainty-quantification,boehm-et-al-2020-prob-autoencoder} showed that the reconstruction of the latent space by posterior inference and by using generative models \cite{adler-et-al-2018-bayesian-inversion,seljiak-et-al-2019-EL2O,wu-et-al-2018-conditional-inference} for a corrupted datum can lead to an optimal image restoration with uncertainty quantification. These methods do not, however, focus on classifying the corrupted datum in the latent space, nor do they use supervised autoencoder structures.
In the field of quantifying uncertainties of classifications, several methods exist. Predominantly, \textsc{Bayesian Neural Networks} (BNNs) \cite{neal-et-al-1995-BNN} and \textsc{Monte Carlo} dropout (MC-dropout) \cite{gal-et-al-2016-dropout} have recently shown success. More recently, \textsc{Evidential Deep Learning} (EDL) \cite{EvidDeepLearning} was introduced as yet another probabilistic method to quantify classification uncertainty. The latter two methods are compared to our method in \Cref{sec:Experiments}.
Finally, various methods to perform image restoration exist in the literature, such as the well-known denoising autoencoder \cite{vincent-et-al-2008-dae}. These conventional methods require prior knowledge of the corruption to be included to the training data.
\subsection{Generative Model and Bayesian Inference with Neural Networks}
The first step of our method is to train a supervised autoencoder. The autoencoder involves the encoding function $f$ (mapping data $\bm{x}\in\mathbb{R}^p$ to the latent space representation with activations $\bm{h}\in\mathbb{R}^z, z\in\mathbb{N}$), as well as the decoding function $g$ (mapping $\bm{h}$ to the data space representation $\hat{\bm{x}}\in\mathbb{R}^p, p\in\mathbb{N}, p\gg z$). The parameters of $f: \mathbb{R}^p \rightarrow \mathbb{R}^z$ and $g: \mathbb{R}^z\rightarrow\mathbb{R}^p$ are optimized via a combination of two loss terms $\mathcal{L}_{gf}$ (representing the reconstruction loss in the data space) and $\mathcal{L}_f$ (representing the classification loss in the latent space): 
\begin{equation}
	\begin{split}
		\mathcal{L}_{\text{SAE}} &=\mathcal{L}_{gf}(g(f(\bm{x})),\bm{x})+\mathcal{L}_f(f(\bm{x})^j,\bm{y}) \\
		&= \mathcal{L}_{gf}(\bm{\hat{x}},\bm{x})+\mathcal{L}_f(\bm{h}^j,\bm{y}),
		\label{eq:loss}
	\end{split}
\end{equation}
where $j$ denotes the number of activations in the latent space $\bm{h}$ that are supervised, i.e., $\bm{h}^j=[\bm{h}_1, \dots, \bm{h}_j]$.
After normalizing all data samples in the range of $[0,1]$, we used the corresponding cross-entropy for each respective loss term to penalize false classifications in the latent space and inaccurate reconstructions in the data space. Note that the loss term $\mathcal{L}_f(\bm{h}^j,\bm{y})$ processes activations from the latent space $\bm{h}^j$ with the $\mathrm{softmax}$-function (i.e., $\mathcal{L}_{\text{SAE}}=\mathcal{L}_{gf}(\bm{\hat{x}},\bm{x})+\mathcal{L}_f(\mathrm{softmax}(\bm{h}^j),\bm{y})$). The $\mathrm{softmax}$-function's output yields values ranging from $[0, 1]$, which can be penalized by one-hot-encoded labels $\bm{y}$. The $\mathrm{softmax}$-function is not included as an activation function in our neural network, where the latent space $\bm{h}$ is activated linearly; see \Cref{sec:Experiments} for details. We minimized the general loss function of \Cref{eq:loss} using the Adam optimizer \cite{kingma-et-al-2014-adam} (test accuracy of $[98.6\%; 89.4\%]$ on the encoding function $f$ with $[$MNIST; Fashion-MNIST$]$). 
Once the training procedure converged, we decoupled the decoding function $g$ from the autoencoder. Without loss of generality, we then used an AWGN model including the nonlinearity $g(\bm{h})$, which additionally involves masking $\bm{m}$ and convolution $\bm{C}$ on $g$: 
\begin{equation}
	\bm{d} = \bm{m}\bm{C}g(\bm{h}) + \bm{n}.
	\label{eq:response-modeling}
\end{equation}
Additive white Gaussian noise, $\bm{n} \in \mathbb{R}^p \sim \mathcal{N}(0, \bm{\Sigma}_n)$, is applied to the decoded latent space signal $g(\bm{h})$, which yields the corrupted data $\bm{d} \in \mathbb{R}^p$. Note that, for the implementation of $\bm{h}=\bm{A\xi}+\bm{\mu}_h$, 
the reparametrization trick \cite{kingma-et-al-2014-aevb} is applied ($\bm{\Sigma}_{\bm{h}} = \mathrm{cov}(f(\bm{X}_{\mathrm{Val}})), \bm{\Sigma}_{\bm{h}}=\bm{A}\bm{A}^\mathrm{T}$,\linebreak $\bm{\Sigma}_{\bm{h}}\in\mathbb{R}^{z\times z}, \bm{\xi}\sim \mathcal{N}(\mathbf{0},\bm{I}), \bm{\xi}\in\mathbb{R}^z$, $\bm{\mu}_h=\mathrm{mean}(\bm{X}_\mathrm{Val}), \bm{\mu}_h \in \mathbb{R}^z$). In addition to AWGN, we included the corruptions of masking $\bm{m}$ and convolutions $\bm{C}$, which are both linear operations.\\
Since we are interested in reconstructing the latent space activation $\bm{h}$ from $\bm{d}$ alongside uncertainty quantification, the goal is to determine the posterior distribution $\mathcal{P}(\bm{h}|\bm{d}) \propto {\mathcal{P}(\bm{d}|\bm{h})\mathcal{P}(\bm{h})}
\label{eq:bayes_law}$.
The log-probability distribution reads
\begin{equation}
	-\ln\mathcal{P}(\bm{h}|\bm{d})=\dfrac{1}{2}\left(\big(\bm{d}-\bm{m}\bm{C}g(\bm{h})\big)^{\mathrm{T}}\bm{\Sigma}_{\bm{n}}^{-1}\left(\bm{d}-\bm{m}\bm{C}g(\bm{h})\right)+\left(\bm{h}^{\mathrm{T}}\bm{\Sigma}_{\bm{h}}^{-1}\bm{h}\right)\vphantom{\big(\bm{h})\big)^{\mathrm{T}}}\right)+\mathrm{const.},
	\label{eq:Information_Hamiltonian}
\end{equation}
where $(\cdot)^{\mathrm{T}}$ denotes the matrix transpose.
Since we are ultimately interested in the analytically intractable mean of $\bm{h},	\left<\bm{h}\right>_{\mathcal{P}(\bm{h}|\bm{d})}=\int \bm{h}\mathcal{P}(\bm{h}|\bm{d}) d\bm{h}
\label{eq:Information_Hamiltonian_h}$, we approximately determined the mean and the variance of $\mathcal{P}(\bm{h}|\bm{d})$ by applying MGVI.
Similar to other variational inference \mbox{methods \cite{kingma-et-al-2014-aevb,kucukelbir-et-al-2017-ADVI}}, MGVI approximates the distribution by a simpler, but tractable distribution from within a variational family, $\mathcal{Q}(\bm{h})$. The parameters of $\mathcal{Q}(\bm{h})$, i.e., mean $\eta$ and covariance $\Delta$, are obtained by minimizing the variational lower bound. The size of a full variational covariance scales quadratically with the number of latent variables. Taking these limitations into account, we employed MGVI, which locally approximates the target distribution using the inverse Fisher metric as an uncertainty estimate around the variational mean $\eta$, which we are optimizing for. The approximation is represented by an ensemble of samples $\mathcal{H}=\{\tilde{\bm{h}}_1, \tilde{\bm{h}}_2, \dots, \tilde{\bm{h}}_n\}$ with $\tilde{\bm{h}} \in \mathbb{R}^z$, which we used for our analysis. $\tilde{\bm{h}}$ refers to the inferred sample. We here call $\overline{\mathcal{H}}$ the posterior mean and $\bm{\delta}_r$ the posterior standard deviation, or the reconstruction uncertainty.

\subsection{Classification and Uncertainty Quantification}
\label{subsec:Classification_Uncertainty}
\begin{figure}[H]
\centering
\input{images/figure3}
	\caption{Accuracy and uncertainty of classifications of data samples of the MNIST dataset at different noise levels (left column, (a)), different masking levels (middle column, (b)), and different convolution levels (right column, (c)) exploiting the supervised latent space structure (top row) with reconstruction uncertainty $\bm{\delta}_r$ and the M-distance $\bm{\delta}_m$ (bottom row) as classifying features. $\mathrm{ACC}_f$ serves as the baseline and is the accuracy of the plain encoding function $f$ classifying corrupted data. $\mathrm{ACC}_g$ corresponds to the accuracy of the method proposed in \Cref{sec:concept_vis}. Additionally, we distinguish between the uncertainties of correct classifications $\bm{\delta}_{r_{\mathrm{True}}}$, $\bm{\delta}_{m_{\mathrm{True}}}$ and of false classifications $\bm{\delta}_{r_{\mathrm{False}}}$, $\bm{\delta}_{m_{\mathrm{False}}}$. The plot was generated with $1000$ test samples for each data point.}
	\label{fig:noise_acc_sae}
\end{figure}
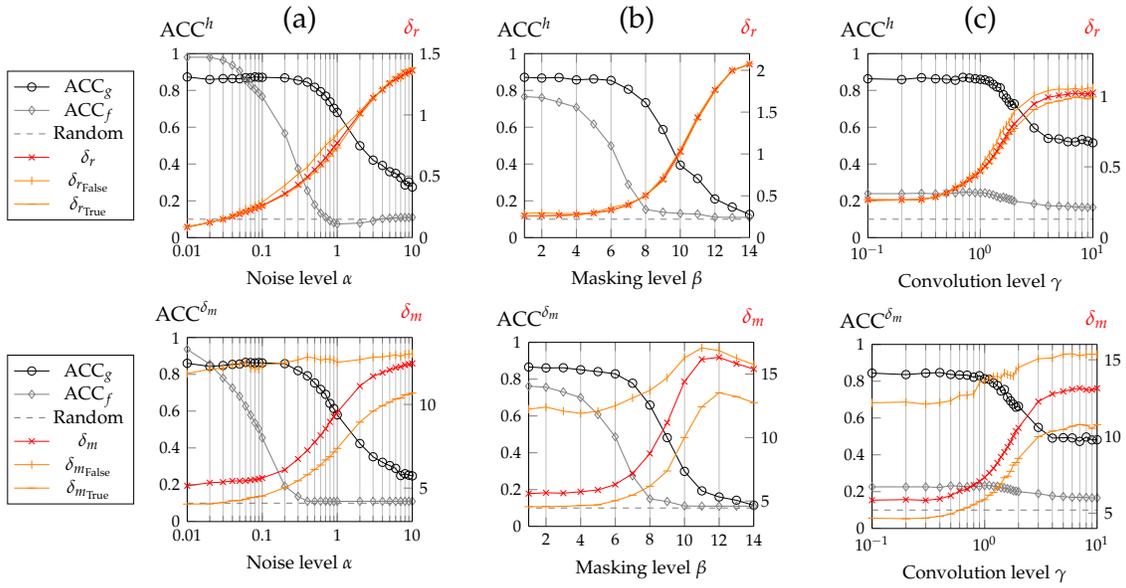

The supervision of the latent space allows us to classify the input $\bm{d}$ in a straightforward manner by evaluating the sample mean and sample standard deviation of the set $\mathcal{H}$. While the sampling mean of the set $\mathrm{mean}(\mathcal{H})=\overline{\mathcal{H}}$ gives the class of the most likely classification, the sampling standard deviation reflects the reconstruction uncertainty $\bm{\delta}_r$ of the latent space posterior distribution. $\bm{\delta}_r$ depends on the type and magnitude of the corruption, as well as the prior probability distribution we included in the channel model (\Cref{eq:Information_Hamiltonian}). We visualized this dependency with various experiments; see \Cref{fig:noise_acc_sae}.

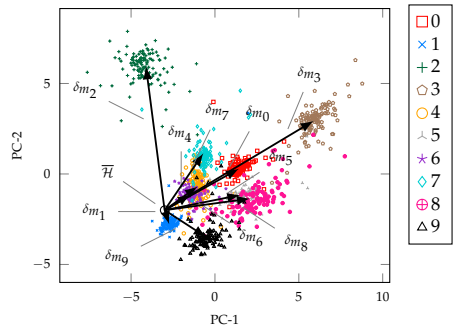
\begin{figure}[H]
\centering

\input{images/figure4}
	\caption{Illustration of the latent space structure of a supervised autoencoder and the M-distance as a classifier based on MNIST. For an arbitrary corrupted datum $\bm{d}$, the inferred posterior mean $\overline{\mathcal{H}}$ in the latent space is marked accordingly. To classify $\overline{\mathcal{H}}$, the M-distance is computed for every cluster in the latent space to obtain $\bm{\delta}_{m_i}$ for all ten classes. The shortest distance $\mathrm{argmin}(\bm{\delta}_m)$ serves as the classification. Samples shown in the figure are transformed to a two-dimensional subspace by principal component analysis. The first two principal components (PC-1, PC-2) are plotted.}
	\label{fig:mahalanobis_graphically}
\end{figure}

Since we are additionally interested in the uncertainty of the model, $\bm{\delta}_m$, we evaluated the M-distance of all samples in $\mathcal{H}$ to every class conditional distribution in the latent space (see arrows in \Cref{fig:mahalanobis_graphically}). We initially determined the parameters of these class conditional distributions by passing the uncorrupted data samples from an independent (i.e., independent of training and testing) dataset $\bm{X}_{\mathrm{Val}}$ (see \Cref{sec:Experiments}) through the encoder $f$. We then evaluated the closest class conditional distribution to a single sample $\tilde{\bm{h}}$, which corresponds to the most likely class. The absolute value of the M-distance to the closest class conditional distribution serves as a measure of the model uncertainty $\bm{\delta}_m$. In this work, all class conditional distributions in the latent space were assumed to follow multivariate Gaussian distributions with covariance $\bm{\Sigma}_i$ and mean $\bm{\mu}_i$. This method is an implementation slightly different from \cite{lee-et-al-2018-mahalanobis-oodd}, where it was shown that the Mahalanobis distance is not only an accurate classifier in this context, but also a reliable out-of-distribution detector reflecting the model uncertainty. Reference \cite{lee-et-al-2018-mahalanobis-oodd} used tied covariance matrices instead of individual covariance matrices for each class conditional distribution, as done in our method. 
\vspace{-6pt}

\section{Experiments}
\label{sec:Experiments}
To validate our method, we conducted several experiments (\label{Code}see for details of implementation and code: \href{https://github.com/pjoppich/corrupted_data_classification}{https://github.com/pjoppich/corrupted\_data\_classification}) on the MNIST \cite{lecunn-1998-mnist-database} and the Fashion-MNIST \cite{xiao-et-al-2017-fmnist} dataset. We evaluated the performance on various corruption types and magnitudes and performed a comparison to MC-dropout \cite{gal-et-al-2016-dropout} and EDL \cite{EvidDeepLearning}. 
The following architecture was used for the supervised autoencoder (we used the same architecture for both datasets):~A~feedforward~neural network was built with~dimen-sions $784^{\{0\}}-512^{\{1\}}-256^{\{2\}}-128^{\{3\}}-10^{\{4\}}-128^{\{5\}}-256^{\{6\}}-512^{\{7\}}-784^{\{8\}}$, where layers $\{0\}-\{2\}$ and $\{4\}-\{7\}$ use the $\mathrm{SeLU}$ activation function \cite{klammbauer-et-al-2017-selu}, layer $\{3\}$ linear,~and layer $\{8\}$ sigmoid activations. Note that, in our case, for simplicity, the number of latent space dimensions $z$ is equal to the number of supervised classes $j$, although $j\leq~z$ holds~generally. We split each dataset into three subsets, $\bm{X}_{\mathrm{Train}}$ ($48\cdot 10^3$ samples, used for training), $\bm{X}_{\mathrm{Test}}$ ($10\cdot 10^3$ samples, used for testing and experiments), and $\bm{X}_{\mathrm{Val}}$ ($12\cdot 10^3$ samples, used for determining $\bm{\Sigma}_{h}$ and ${\bm{\Sigma}_{\mathcal{C}_1}} \dots {\bm{\Sigma}_{\mathcal{C}_K}}$).
We used the MGVI implementation of NIFTy \cite{selig-et-al-nifty} to perform the inference\footnotemark[3]. 
\subsection{Classification}
We visualize experiments (1)--(3) in \Cref{fig:noise_acc_sae}. In the first experiment (1), we classified data from an independent test set of the MNIST-dataset corrupted by different noise levels with the proposed method. We denote $\alpha$ as the noise level of $n$, $n \sim \mathcal{N}(0, \alpha)$. We compared the accuracy of our method to the baseline of processing corrupted data through the encoder of the pretrained autoencoder. We show that we significantly improved the accuracy of classifying corrupted data in comparison to the straightforward classification by $f(\bm{d})$. 
For the second experiment (2), we used the same data samples as for (1) with the exception that we now additionally corrupted the data with window masking at a constant noise level of $\alpha=0.1$. Again, we compared the accuracy of our method to the baseline of processing the same data samples through the encoder.
In the third experiment (3), we corrupted the data by convolving them with a Gaussian blur kernel with a filter size of $7~\times~7$ and different magnitudes $\gamma$ at a constant noise level of $\alpha=0.1$. 

Experiments (1), (2), and (3) led to the following conclusions:
\vspace{-0.1cm}
\begin{itemize}
 \item The reconstruction uncertainty $\bm{\delta}_{r_{\mathrm{True}}}$ of correct classifications is approximately equivalent to the $\bm{\delta}_{r_{\mathrm{False}}}$ of wrong classifications. This behavior indicates that the correctness of the classification does not influence the reconstruction uncertainty $\bm{\delta}_{r}$, showing evidence that $\bm{\delta}_{r}$ is independent of the model uncertainty $\bm{\delta}_m$.
 \item As opposed to $\bm{\delta}_r$, the model uncertainty $\bm{\delta}_m$ strongly depends on the correct/wrong classification of the corrupted datum: $\bm{\delta}_m$ is significantly and consistently higher for false classifications than for true classifications. This characteristic sets the basis for a statistical ``lie detector'' (see \Cref{subsec:det_false_class}) of classification. Fields of application could be the validation of neural networks in, e.g., medical imaging and other safety-critical applications.
 \item Classifying corrupted data through the decoder (see $\mathrm{ACC}_g$ in \Cref{fig:noise_acc_sae}) (rather than the encoder (see $\mathrm{ACC}_f$)), with a suitable channel model considering the corruption, significantly improved the model's accuracy without the necessity of retraining the autoencoder. Especially for high levels of all corruption types, the accuracy of the model notably improved. Corruption by convolution had catastrophic consequences for classifying data in a straightforward manner through the encoder $f$, while this type of corruption seemed to only have a minor impact on our method. 
 \item Both uncertainties $\bm{\delta}_r$ and $\bm{\delta}_m$ rose with increasing levels of corruption.
\end{itemize}
\subsection{Detection of False Classifications}
\label{subsec:det_false_class}
Finally, in experiment (4) (see \Cref{fig:roc}), we validated the model uncertainty of our method by introducing the Uncertainty-based Receiver Operating Characteristics (U-ROC) curve of detecting false classifications with the M-distance. We evaluated the binary classification task of the two classes ``\textit{The neural network correctly classifies a corrupted datum}'' (\textsc{Positive Class}) and ``\textit{The neural network falsely classifies a corrupted datum}'' (\textsc{Negative Class}). Based on the model uncertainty of our method, we aimed to predict the two classes without further knowledge, providing the initially proposed ``lie detector''. The U-ROC curve was built from the \textsc{True Positive Rate} and the \textsc{False Positive Rate}.
We compared our U-ROC curve with the U-ROC curve of the MC-dropout method \cite{gal-et-al-2016-dropout} and with the U-ROC curve of EDL \cite{EvidDeepLearning}, feeding all methods with the identical input of a datum corrupted by noise at $\alpha=[0.1, 0.5, 1.0]$.
We made the following conclusions from experiment (4), \Cref{fig:roc}:
\vspace{-0.1cm}
\begin{itemize}
 \item Our method seemed to outperform MC-dropout and EDL to detect false classifications given the same data samples at the input for $\alpha=0.1$ and $\alpha=0.5$. One reason for this might be that the M-distance serves as a reliable out-of-distribution detector, exploiting the inherent latent space structure of uncorrupted data as a reference, as opposed to MC-dropout and EDL. For $\alpha=1.0$, both EDL and our method outperformed MC-dropout, while the Area Under the Curve (AUC) of EDL was largest. Here, it should be noted that EDL cannot classify the corrupted data at this noise level (accuracy: $8.9\%$), resulting in only few samples to test the cases of \textsc{True Positives} and \textsc{False Positives}.
 \item All three methods provided reliable results for detecting false classifications for low noise levels.
The model uncertainty $\bm{\delta}_m$ truly reflects the confidence of the classification, i.e., a high value of $\bm{\delta}_m$ correlates empirically with a higher probability of false classification.
\item The U-ROC curve combined with the accuracy indicates that EDL seemed to overestimate uncertainties, leading to a very robust U-ROC curve for high noise levels, but simultaneously leading to a severe drop in the accuracy in the presence of data corruption. We observed comparable results on F-MNIST data.
\end{itemize}
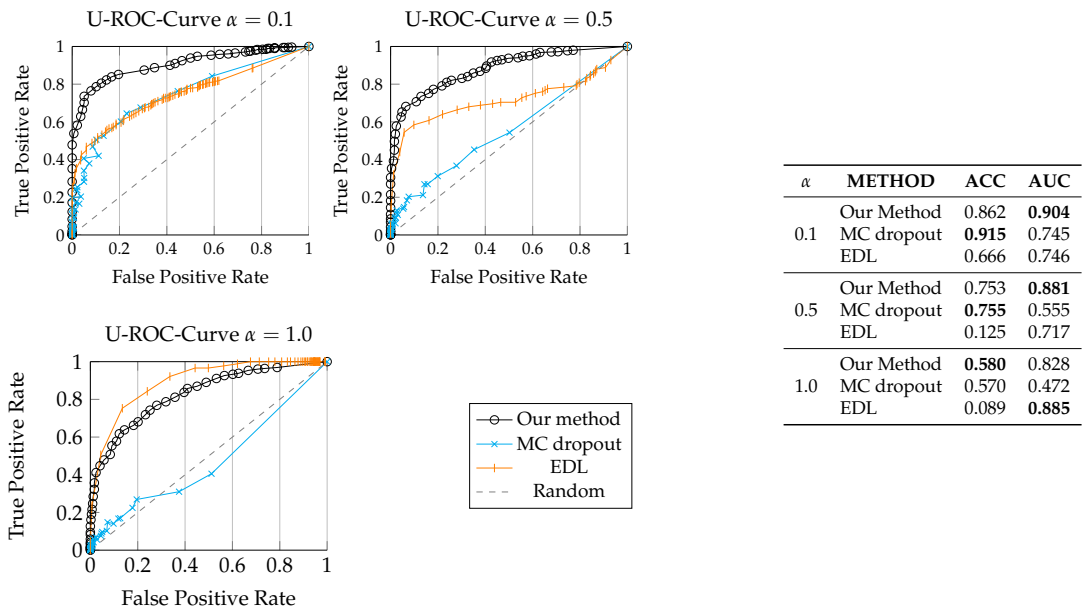
\begin{figure}[H]
\input{images/figure5}

	\caption{U-ROC of the proposed identifier of false classifications for different noise levels $\alpha$ of our method in comparison with MC-dropout and with EDL. In this experiment, the formulation of the, e.g., ``\textsc{True Negative}'' case would be: \textit{based on the uncertainty value, the sample is correctly detected as a false classification}---the ``lie detector'' works. Samples are taken from the MNIST-dataset. Top left: corrupted datum at $\alpha=0.1$. Bottom left: corrupted datum at $\alpha=1.0$. The irregularity in the U-ROC curve of the dropout model is due to the stochastic nature of MC-dropout. Bottom right: Evaluation of Accuracy ($\mathrm{ACC}$) for all given noise values $\alpha$ and the $\mathrm{AUC}$ for all U-ROCs.}
	\label{fig:roc}
\end{figure}
\section{Requirements and Summary}
Our proposed methodology to classify a corrupted datum $\bm{d}$ including uncertainty quantification requires the following inputs in addition to $\bm{d}$:
\vspace{-0.1cm}
\begin{itemize}
 \item $\bm{m}, \bm{C}$: Without loss of generality, here, we assumed corruption by masking and convolution represented by $\bm{m}$ and $\bm{C}$ in the AWGN channel model, as depicted in \Cref{eq:response-modeling}. Here, $\bm{C}$ can in real-world applications often be derived from the image processing system in use. Algorithms to detect possibly occluding objects (represented by masking $\bm{m}$) were given by, e.g., \cite{li2013modeling}.
 \item $\bm{\Sigma}_{\bm{n}}$: Noise covariance matrix. AWGN with $n \sim \mathcal{N}(0, \Sigma_n)$ and applied additively to the data $\bm{d}$. Among others, the methodology published by \cite{liu2012noise} enables the derivation of $\bm{\Sigma}_{\bm{n}}$ given noisy data $\bm{d}$.
 \item $\bm{\Sigma}_{\bm{h}}$: Sampling covariance matrix of all (uncorrupted) latent space activations processed by the encoding function $f$. We used the assumption that an autoencoder can represent an inherent, lower-dimensional structure of the data in its latent space and assumed this sub-dimensional structure to sufficiently follow a multivariate Gaussian probability distribution. 
\end{itemize}

In summary, our approach was able to classify heavily corrupted data with parametric classifiers. The method does not require corrupted data for training. As we built our procedure on a probabilistic architecture, we quantified the classification and the model uncertainties, allowing for a reliable detection of false classifications. We see our method as a highly flexible and robust framework that can be applied to \textit{any} generative neural network to improve performance on corrupted data significantly. If the generative neural network comes with a supervised encoded space, it can classify the data directly. We showed that the M-distance can independently be used to classify data. The limitations of our method include that the corruption type needs to be modeled, as well as there is a higher computational cost than MC-dropout and EDL (mainly due to the approximation of the posterior probability distribution; Step 4 in Section \ref{sec:concept_vis}).

\vspace{6pt}

\authorcontributions{Conceptualization, P.J., S.D. and O.D.C.; methodology, P.J., S.D. and O.D.C.; software, P.J. and J.K.; validation, P.J. and S.D.; formal analysis, P.J., S.D. and O.D.C., J.K. and W.U.; investigation, P.J.; resources, n.a.; data curation, P.J.; writing---original draft preparation, P.J., S.D. and O.D.C.; writing---review and editing, P.J., S.D., O.D.C., J.K. and W.U.; visualization, P.J.; supervision, W.U.; project administration, S.D., O.D.C. and W.U.; funding acquisition, n.a. All authors have read and agreed to the published version of the manuscript.}

\funding{This research received no external funding.}

\institutionalreview{Not applicable.}

\informedconsent{Not applicable.}


\dataavailability{Data publicly available, http://yann.lecun.com/exdb/mnist/ (MNIST) and https://github.com/zalandoresearch/fashion-mnist (F-MNIST).}

\acknowledgments{The authors thank T.A. Enßlin for valuable discussions and for supporting the implementation of the Bayesian inference methods. Furthermore, we thank AUDI AG and the Technical University of Munich for providing an optimal research environment. Jakob Knollmüller acknowledges the financial support by the Excellence Cluster ORIGINS, which is funded by the Deutsche Forschungsgemeinschaft (DFG, German Research Foundation) under Germany's Excellence Strategy---EXC-2094-390783311.}

\conflictsofinterest{The funders had no role in the design of the study; in the collection, analyses, or interpretation of data; in the writing of the manuscript, or in the decision to publish the~results.}
\begin{adjustwidth}{-\extralength}{0cm}

\reftitle{References}

\end{adjustwidth}
\end{document}

%% file: images/figure1.tex
\begin{tikzpicture}

\node(0){\includegraphics[scale=0.25]{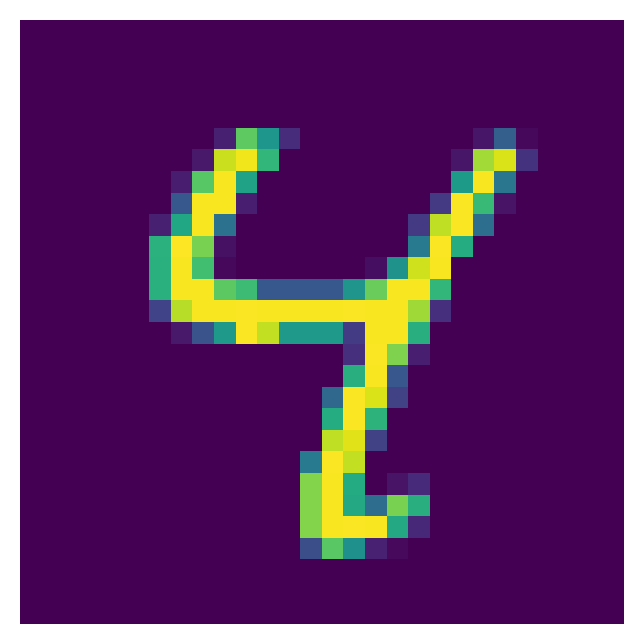}};
\node[right=0.25cm of 0](1){\includegraphics[scale=0.25]{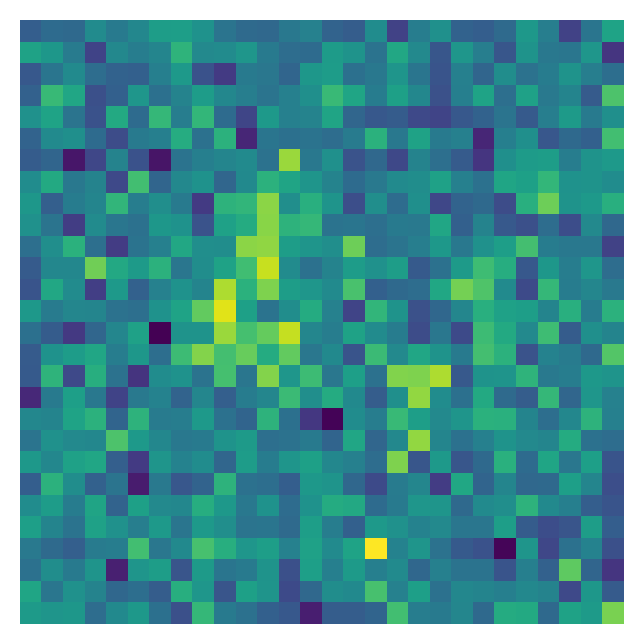}};
\node[xshift=2.5cm, yshift=0.4cm](bars1) at (1.east) {\begin{tikzpicture}

\begin{axis}[width=3.5cm, height=3.5cm,ymin=-6, ymax=7.5,align =center, title={\scriptsize{$\overline{\mathcal{H}}\pm\bm{\delta_r}$ \& $f(\bm{x})$}},bar width=3pt,
ticklabel style = {font=\tiny},label style={font=\tiny}, xmajorticks=false, minor xtick={0,1,2,3,4,5,6,7,8,9}]
    \addplot[fill=gray,
        ybar,
        error bars/.cd,
            y dir=both, y explicit,
            error bar style={line width=0.75pt},
    error mark options={
      mark size=0pt,
      rotate=90,
      line width=0.5pt,}
    ] coordinates {
( 0 , -1.333790594849042 ) +- ( 0 , 0.3233276660861007 )
( 1 , -1.1111370552955604 ) +- ( 1 , 0.2225506278536135 )
( 2 , -0.9910260657828902 ) +- ( 2 , 0.41685151554656413 )
( 3 , -1.1539068430125368 ) +- ( 3 , 0.4447458254221104 )
( 5 , -0.29441899129954535 ) +- ( 5 , 0.37324412770211873 )
( 6 , -0.9489737369890273 ) +- ( 6 , 0.2104434266205051 )
( 7 , -0.18124802162347353 ) +- ( 7 , 0.5442769239194797 )
( 8 , 1.612331841522809 ) +- ( 8 , 0.8466116321569718 )
( 9 , -0.4826476718967445 ) +- ( 9 , 0.5321798306311084 )
};

    \addplot[fill=orange, ybar, error bars/.cd, y dir=both, y explicit,
            error bar style={line width=0.75pt},
    error mark options={
      mark size=0pt,
      rotate=90,
      line width=0.5pt}]
    coordinates{
( 4 , 5.206786750464037 ) +- ( 4 , 1.7419255623674585 )
};
\addplot[fill=white,opacity=0.35,
	bar width=2pt,
        ybar,
       ] coordinates {
( 0 , -1.2548176 )
( 1 , -0.89296407 )
( 2 , -0.7592658 )
( 3 , -1.0203555 )
( 4 , 6.2007537 )
( 5 , -0.3555865 )
( 6 , -1.0464954 )
( 7 , -0.37164295 )
( 8 , 0.22147377 )
( 9 , -0.2920838 )
};
\end{axis}
\end{tikzpicture}
\begin{tikzpicture} 
\begin{axis}[width=3.5cm, height=3.5cm,ymin=-2, ymax=28, title={\scriptsize{$\bm{\delta_m} \pm \bm{\delta_r}$}},bar width=3pt,
ticklabel style = {font=\tiny},xmajorticks=false, minor xtick={0,1,2,3,4,5,6,7,8,9}]
    \addplot[fill=gray,
        ybar,
        error bars/.cd,
            y dir=both, y explicit,
            error bar style={line width=0.75pt},
    error mark options={
      mark size=0pt,
      rotate=90,
      line width=0.5pt}
    ] coordinates {
( 0 , 20.060900428974247 ) +- ( 0 , 2.481000554583958 )
( 1 , 20.302899200116165 ) +- ( 1 , 3.931952257097683 )
( 2 , 13.478442920276086 ) +- ( 2 , 2.4467014087870234 )
( 3 , 16.663533187713163 ) +- ( 3 , 3.3978423168627865 )
( 5 , 15.009544905231017 ) +- ( 5 , 2.847201424411398 )
( 6 , 21.421548279312482 ) +- ( 6 , 4.588945759841051 )
( 7 , 17.030865720530574 ) +- ( 7 , 3.5992408442589046 )
( 8 , 15.150060199892284 ) +- ( 8 , 3.5013285936482794 )
( 9 , 12.692934176045009 ) +- ( 9 , 2.425891296322672 )
    };
    \addplot[fill=orange, ybar, error bars/.cd, y dir=both, y explicit,
            error bar style={line width=0.75pt},
    error mark options={
      mark size=0pt,
      rotate=90,
      line width=0.5pt}]
    coordinates{
( 4 , 5.288998946438056 ) +- ( 4 , 1.1408347018587885 )
};
\end{axis}
\end{tikzpicture}};


\node[below=0.1cm of 0](0a){\includegraphics[scale=0.25]{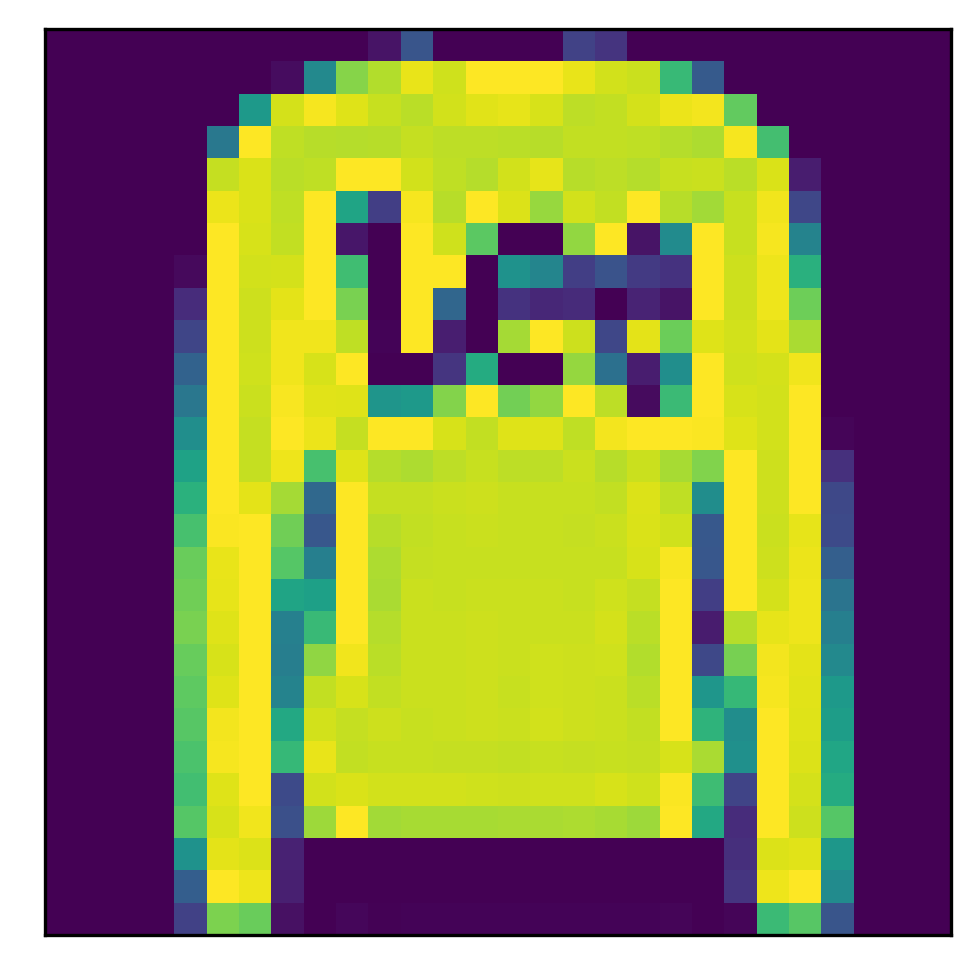}};
\node[right=0.25cm of 0a](1a){\includegraphics[scale=0.25]{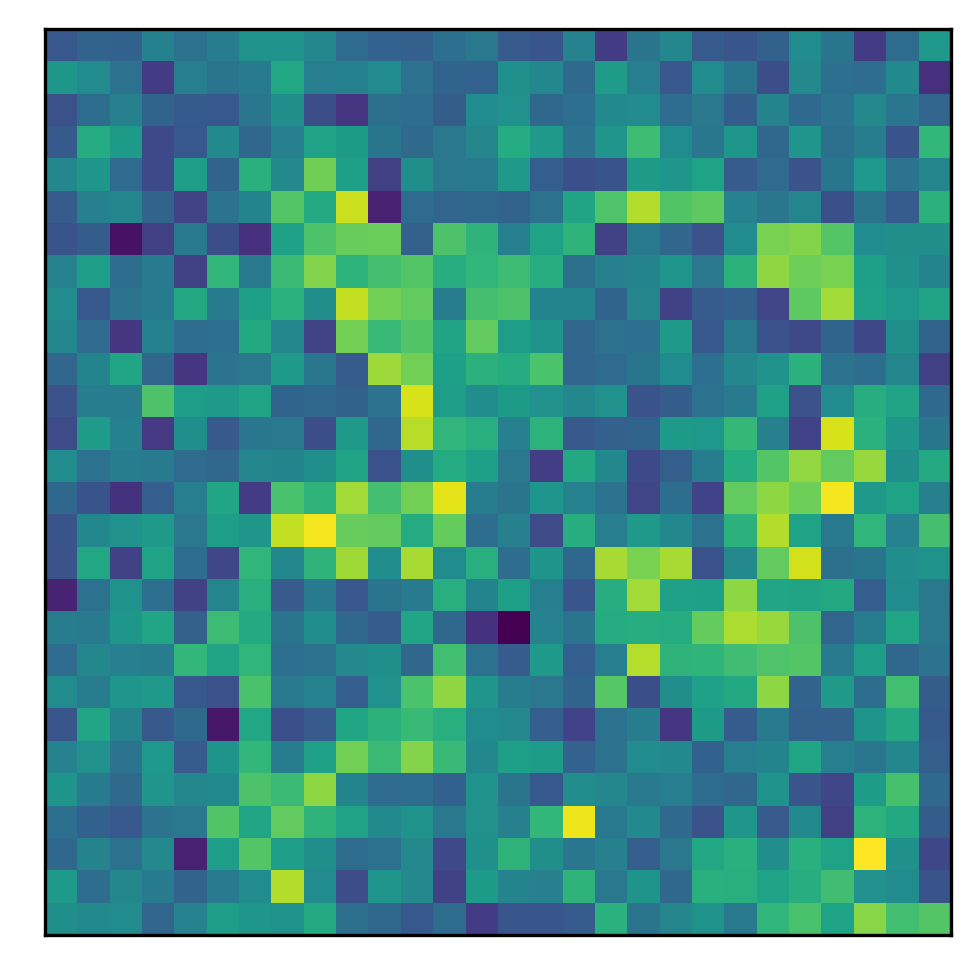}};
\node[xshift=2.5cm, yshift=-0.15cm](bars2)at (1a.east){\begin{tikzpicture} 
\begin{axis}[width=3.5cm, height=3.5cm,ymin=-8, ymax=13,bar width=3pt,
ticklabel style = {font=\tiny}, label style={font=\tiny}, minor xtick={0,1,2,3,4,5,6,7,8,9},xtick={0,1,2,3,4,5,6,7,8,9}]
    \addplot[fill=gray,
        ybar,
        error bars/.cd,
            y dir=both, y explicit,
            error bar style={line width=0.75pt},
    error mark options={
      mark size=0pt,
      rotate=90,
      line width=0.5pt}
    ] coordinates {
(0, 0.9074392244183155)+-(0,1.2897490155704396)
(1, -4.977045835125597)+-(0,1.6892428130341155)
(3, -6.169149900227663)+-(0,1.4906571291241688)
(4, 7.0016367542685)+-(0,1.9787407434511237)
(5, 1.0576525645117487)+-(0,0.8607711571438896)
(6, 9.170315358639554)+-(0,1.9154243885163944)
(7, -5.311896177195083)+-(0,1.9562738457380366)
(8, -0.9274058253085375)+-(0,1.540772192430144)
(9, 0.24850070002522637)+-(0,1.2584509020932937)
};
    \addplot[fill=orange,
        ybar,
        error bars/.cd,
            y dir=both, y explicit,
            error bar style={line width=0.75pt},
    error mark options={
      mark size=0pt,
      rotate=90,
      line width=0.5pt}
]
    coordinates{
(2, 9.843916010420887)+-(0,2.4146014584836357)
};
\addplot[fill=white,opacity=0.35,
	bar width=2pt,
        ybar,
       ] coordinates {
    (0, -0.12554067)
(1,  -6.1688843)
(2,  8.908215)
(3, -4.1438165 )
(4, 5.3111887)
(5,  -0.625254 )
(6,  7.619137)
(7,  -6.4036336 )
(8,  -2.3990023 )
(9,  -0.45938364)
};

\end{axis}
\end{tikzpicture}%
    ~%
\begin{tikzpicture} 
\begin{axis}[width=3.5cm, height=3.5cm,ymin=-1, ymax=13,bar width=3pt,
ticklabel style = {font=\tiny},label style={font=\tiny}, minor xtick={0,1,2,3,4,5,6,7,8,9}, xtick={0,1,2,3,4,5,6,7,8,9}]
    \addplot[fill=gray,
        ybar,
        error bars/.cd,
            y dir=both, y explicit,
            error bar style={line width=0.75pt},
    error mark options={
      mark size=0pt,
      rotate=90,
      line width=0.5pt}
    ] coordinates {
(0, 8.707195023162308)+-(0,1.4977375834377287)
(1, 9.176380978858075)+-(0,1.1452570537905868)
(2, 4.814811702723542)+-(0,1.1998797423448213)
(3, 6.930808346406654)+-(0,0.7820257299737633)
(5, 6.569816516811618)+-(0,0.8812136038399144)
(6, 5.045057432101427)+-(0,1.106191080862397)
(7, 10.058354320681083)+-(0,1.2246883856101713)
(8, 6.8303660543352045)+-(0,0.9496418695769726)
(9, 8.664000429789452)+-(0,1.4414580960060097)
    };
    \addplot[fill=orange,
        ybar,
        error bars/.cd,
            y dir=both, y explicit,
            error bar style={line width=0.75pt},
    error mark options={
      mark size=0pt,
      rotate=90,
      line width=0.5pt}
]
    coordinates{
(4, 4.507322901808305)+-(0,0.9295378337849619)
};
\end{axis}
\end{tikzpicture}};
  \begin{scope}[shift={(3,0)}]
  \draw[decorate, decoration={brace, amplitude=1ex, raise=1ex}](1.2, 1.6) -- (5.8, 1.6) node[pos=.5, above=2.5ex, align=center] {\scriptsize{\textsc{Latent-Space}:}\\ \scriptsize{Classification and Uncertainty Quantification}};
    \end{scope}
\node[right=0.25cm of bars1, yshift=-0.4cm](2){\includegraphics[scale=0.25]{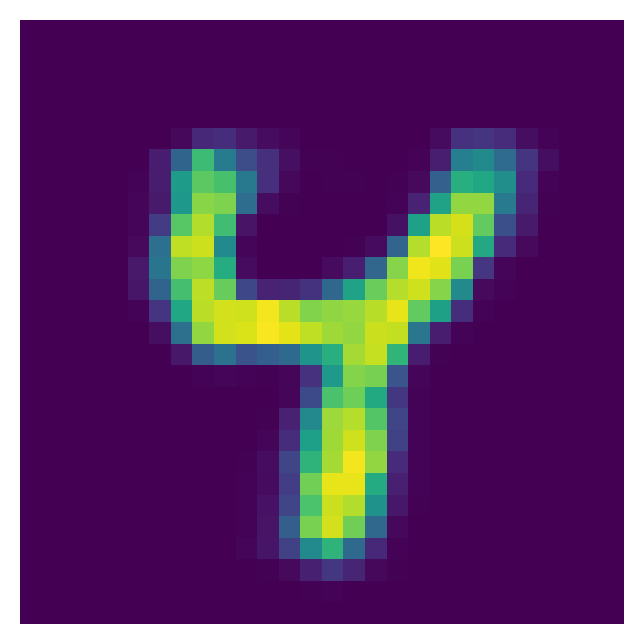}};
\node[above=0.05cm of 2, text width=3cm, align=center](2a){\scriptsize{\textsc{Reconstruction}\\$g(\overline{\mathcal{H}})$}};
\node[right=0.19cm of bars2, yshift=0.15cm](2a){\includegraphics[scale=0.25]{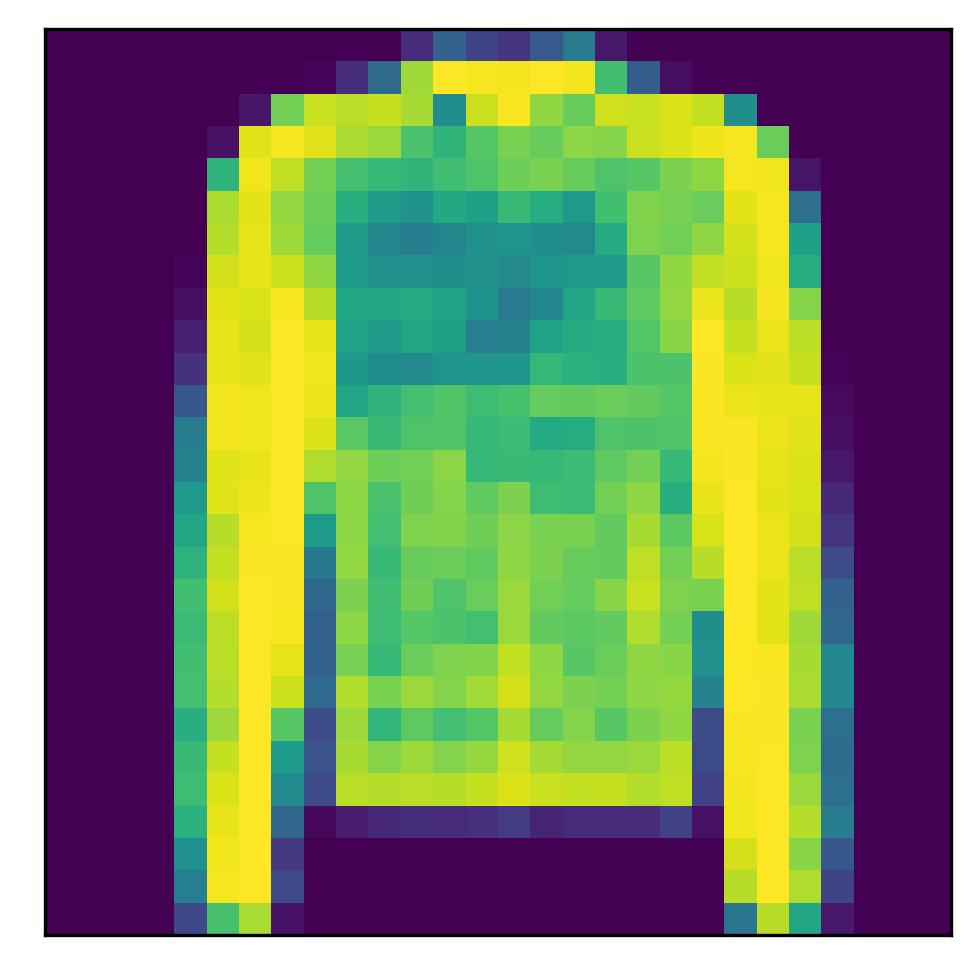}};
\node[above=0.05cm of 0, text width=3cm, align=center](0a){\scriptsize{\textsc{Ground-Truth}\\ $\vphantom{\bm{mC}}\bm{x}$}};
\node[above=0.05cm of 1, text width=3cm, align=center](1a){\scriptsize{\textsc{Corrupted Datum}\\ $\bm{d}=\bm{mC}\bm{x}+\bm{n}$}};
\end{tikzpicture}

%% file: images/figure2.tex
		  \begin{tikzpicture}[scale=0.6]
  \draw[ultra thick, color=gray] (0, 0) -- (0, 3.92);
    \draw[dashed, -latex, color=gray] (0, 3.92) --(1, 3.2);
    \draw[dashed, -latex, color=gray] (0, 0) --(1,0.64);
    \node at (0, 4.5) {\textcolor{gray}{\tiny{$[784]$}}};
    \node at (-0.2, 1.96) (x) {\textcolor{gray}{\tiny{$\bm{x}$}}};
    
    \draw[ultra thick, color=gray] (1,0.64) -- (1,3.2);
    \draw[dashed, -latex, color=gray] (1,3.2) --(2, 2.6);
    \draw[dashed, -latex, color=gray] (1,0.64) --(2,1.32);
    \node at (1, 4.0) {\textcolor{gray}{\tiny{$[512]$}}};
    
    \draw[ultra thick, color=gray] (2,1.32) -- (2, 2.6);
    \draw[dashed, -latex, color=gray] (2,2.6) --(3, 2.28);
    \draw[dashed, -latex, color=gray] (2,1.32) --(3,1.64);
    \node at (2, 3.5) {\textcolor{gray}{\tiny{$[256]$}}};
    
    \draw[ultra thick, color=gray](3, 1.64) -- (3, 2.28);
    \draw[dashed, -latex, color=gray] (3,2.28) --(4, 1.9858);
    \draw[dashed, -latex, color=gray] (3,1.64) --(4,1.935);
    \node at (3, 3.2) {\textcolor{gray}{\tiny{$[128]$}}};
    
    \draw[ultra thick] (4,1.935) -- (4,1.985);
    \draw[dashed, -latex] (4,1.985) --(5, 2.28);
    \draw[dashed, -latex] (4,1.935) --(5,1.64);
    \node at (4, 3.0) {\tiny{$[10]$}};
    \node at (4, 1.6) {\tiny{$\bm{h}$}};
    
    \draw[ultra thick] (5,1.64) -- (5,2.28);
    \draw[dashed, -latex] (5,2.28) --(6, 2.6);
    \draw[dashed, -latex] (5,1.64) --(6,1.32);
    \node at (5, 3.2) {\tiny{$[128]$}};    
        
    \draw[ultra thick] (6,1.32) -- (6, 2.6);
    \draw[dashed, -latex] (6,2.6) --(7, 3.2);
    \draw[dashed, -latex] (6,1.32) --(7,0.64);
    \node at (6, 3.5) {\tiny{$[256]$}};    
        
    \draw[ultra thick](7, 0.64) -- (7, 3.2);
    \draw[dashed, -latex] (7,3.2) --(8, 3.92);
    \draw[dashed, -latex] (7,0.64) --(8,0);
    \node at (7, 4.0) {\tiny{$[512]$}};    
    
    \draw[ultra thick] (8,0) -- (8,3.92);
    \node at (8, 4.5) {\tiny{$[784]$}};
    \node at (8.5, 1.96)(hat_x) {\tiny{$g(\bm{h})$}};
    

    \draw[ultra thick](10, 0) -- (10, 3.92);
    \node at (10, 4.5) {\tiny{$[784]$}};
    \node at (10.25, 1.96)(hat_x) {\tiny{$\bm{n}$}};
    \node at (9.3, 1.96)(a) {\tiny{$+$}};
    \node at (10.8, 1.96)(a) {\tiny{$=$}};
    
    \node at (11.5, 4.5) {\tiny{$[784]$}};
    \draw[ultra thick](11.5, 0) -- (11.5, 3.92);
	 \node at (11.75, 1.96)(hat_x) {\tiny{$\bm{d}$}};

\def\centerx{2}
\def\centery{-1}
	\draw[thick, dashed](3.6, -0.1) -- (3.6, 4.9) -- (9,4.9) -- (9, -0.1) -- (3.65, -0.1);
	\node[circle,draw,outer sep=0pt,inner sep=0.1ex] at (-0.5, 5.25)(a) {1};
	\node[circle,draw,outer sep=0pt,inner sep=0.1ex] at (8.7, 5.25)(a) {2};
	
	\draw [decorate,decoration={brace,amplitude=5pt,mirror,raise=4pt},yshift=0pt, thick]
(8,-0.3) -- (11.75,-0.3) node [black,midway,yshift=-0.8cm] {};

		\node[circle,draw,outer sep=0pt,inner sep=0.1ex] at (9.875, -1.2)(a) {3};
			\draw[-latex, thick](12, 1.96)--(12.5, 1.96)--(12.5, -1.8)--(4.6, -1.8);

	 \node at (3, -1.8) {\begin{axis}[width=2.8cm, height=2.8cm,    hide axis,colormap = {whiteblack}{color(0cm)  = (white);color(1cm) = (black)}
]
		
    \addplot3[surf,
    domain=-2:6,
    domain y=-5:3, 
] 
        {exp(-( (x-\centerx)^2 + (y-\centery)^2)/3 )};
    \end{axis}};
	
		\node[circle,draw,outer sep=0pt,inner sep=0.1ex] at (12.85, 0)(a) {4};
	\node at (4, -1.95){\tiny{$\mathcal{P}(\bm{h}|\bm{d})$}};
	
	\draw[-latex, thick](4, -2.2)--(4, -2.8);
	\node at (4, -3) {\tiny{$\mathcal{H}=\{\tilde{\bm{h}}_1, \tilde{\bm{h}}_2, \hdots, \tilde{\bm{h}}_n\}$}};
	\node at (9, -2.1){\tiny{\textsc{Variational Inference}}};
  \end{tikzpicture}

%% file: images/figure3.tex
	\centering
	\begin{minipage}[t]{0.3\textwidth}
		\pgfplotsset{compat=1.3,small}
\begin{tikzpicture}[baseline, scale=0.78]
\begin{axis}[scale=0.78,
xmode=log,
            log ticks with fixed point,
            axis y line*=left,
            ymin=0, ymax=1,
            xmin=0.01, xmax=10,
            xmajorgrids,
            xminorgrids,
            	title=\Large{(a)},
            xlabel=Noise level $\alpha$,
            ylabel=$\mathrm{ACC}^{h}$,every axis y label/.style={
    at={(ticklabel* cs:1.05)},
    anchor=south,
},]
\addplot[black, mark=o] coordinates {
(0.01,0.873)
(0.02,0.859)
(0.03,0.864)
(0.04,0.864)
(0.05,0.863)
(0.06,0.869)
(0.07,0.869)
(0.08,0.873)
(0.09,0.871)
(0.1,0.871)
(0.2,0.868)
(0.3,0.855)
(0.4,0.845)
(0.5,0.817)
(0.6,0.793)
(0.7,0.769)
(0.8,0.738)
(0.9,0.703)
(1,0.68)
(2,0.499)
(3,0.421)
(4,0.392)
(5,0.358)
(6,0.349)	
(7,0.326)
(8,0.286)
(9,0.302)
(10, 0.275)
    };
  \label{plot_one_top}

    \addplot[gray, mark=diamond] coordinates {   
(0.010000,0.980000)
(0.020000,0.980000)
(0.030000,0.964000)
(0.040000,0.937000)
(0.050000,0.906000)
(0.060000,0.856000)
(0.070000,0.828000)
(0.080000,0.807000)
(0.090000,0.786000)
(0.100000,0.766000)
(0.200000,0.567000)
(0.300000,0.375000)
(0.400000,0.255000)
(0.500000,0.179000)
(0.600000,0.133000)
(0.700000,0.111000)
(0.800000,0.092000)
(0.900000,0.081000)
(1.000000,0.074000)
(2.000000,0.079000)
(3.000000,0.090000)
(4.000000,0.103000)
(5.000000,0.106000)
(6.000000,0.107000)
(7.000000,0.109000)
(8.000000,0.109000)
(9.000000,0.109000)
(10.000000,0.110000)
};
\addplot[gray,smooth, mark=none, dashed] coordinates {
(0.01,0.1)
(10,0.1)
    };
\end{axis}  
\begin{axis}[scale=0.78,xmode=log,
            log ticks with fixed point,
            axis y line*=right,
            ymin=0, ymax=1.5,
            xmin=0.01, xmax=10,
            axis x line=none,
            ylabel=\textcolor{red}{${\delta}_r$},every axis y label/.style={
    at={(ticklabel* cs:1.05)},
    anchor=south,
},
            legend style={at={(-0.8,0.5)},anchor=west},
            legend style={nodes={scale=0.9, transform shape}}]
                        \addlegendimage{mark=o}
\addlegendentry{$\mathrm{ACC}_g$}

\addlegendimage{mark=diamond, gray}
\addlegendentry{$\mathrm{ACC}_f$}
\addlegendimage{mark=none, gray, dashed}
\addlegendentry{Random}
\addplot[red, mark=x] coordinates {
(0.01,0.08727)
(0.02,0.12374)
(0.03,0.14935)
(0.04,0.17221)
(0.05,0.1916)
(0.06,0.20772)
(0.07,0.22236)
(0.08,0.23745)
(0.09,0.2486)
(0.1,0.25998)
(0.2,0.35764)
(0.3,0.43056)
(0.4,0.49578)
(0.5,0.55301)	
(0.6,0.60311)
(0.7,0.65036)
(0.8,0.69613)
(0.9,0.73638)
(1,0.7702)
(2,1.01175)
(3,1.13936)
(4,1.2064)
(5,1.258995)
(6,1.29306)
(7,1.31226)
(8,1.33906)	
(9,1.34993)
(10, 1.3636862969374017)
    };

    \addplot[orange, mark=|] coordinates {
(0.01, 0.08312750807795001)
(0.02, 0.12258390671734395)
(0.03, 0.15230219240964676)
(0.04, 0.1791954201983154)
(0.05, 0.20930656253720917)
(0.06, 0.23186168561335266)
(0.07, 0.23957667070783906)
(0.08, 0.26000719569976133)
(0.09, 0.27048951796661536)
(0.1, 0.2885218280993516)
(0.2, 0.42531845486211456)
(0.3, 0.5182981070573495)
(0.4, 0.5737026506983514)
(0.5, 0.6488149945407047)
(0.6, 0.7033399479860463)
(0.7, 0.7521623856642772)
(0.8, 0.7847078027733624)
(0.9, 0.8060964414208853)
(1, 0.8517521277375716)
(2, 1.02754821137981)
(3, 1.1480988912729748)
(4, 1.2025423947827127)
(5, 1.2527494527313991)
(6, 1.2839360227174272)
(7, 1.3023477725567925)
(8, 1.3284738982748039)
(9, 1.3405075496755046)
(10, 1.356294859733464)
    };
\label{std_false_top}
\addplot[orange, mark=-] coordinates {
(0.01, 0.087872182442168)
(0.02, 0.1239365804337927)
(0.03, 0.1488914081497254)
(0.04, 0.1711067455335288)
(0.05, 0.1888000326283356)
(0.06, 0.20408075107891424)
(0.07, 0.219772318452004)
(0.08, 0.2341750218389143)
(0.09, 0.24535910519290544)
(0.1, 0.25575482613386946)
(0.2, 0.34735722722781026)
(0.3, 0.41568919937417803)
(0.4, 0.48148892615832484)
(0.5, 0.5315496833528716)
(0.6, 0.5769504551361601)
(0.7, 0.6209517435930185)
(0.8, 0.6646869743177126)
(0.9, 0.7066369512123996)
(1, 0.7318193325860743)
(2, 0.99588524818257)
(3, 1.1256970643777897)
(4, 1.2123811956639095)
(5, 1.2701991549674116)
(6, 1.310078049960897)
(7, 1.332772177784936)
(8, 1.3654990848510238)
(9, 1.371724091227549)
(10, 1.3831728132023284)
    };
\label{std_true_top}

\addlegendimage{/pgfplots/refstyle=plot_one_top}\addlegendentry{${\delta}_r$}
\addlegendimage{/pgfplots/refstyle=std_true_top}\addlegendentry{${\delta}_{r_{\mathrm{False}}}$}
\addlegendimage{/pgfplots/refstyle=std_false_top}\addlegendentry{${\delta}_{r_{\mathrm{True}}}$}

  \end{axis}
\end{tikzpicture}\\
\begin{tikzpicture}[baseline, scale=0.78]
\begin{axis}[scale=0.78,xmode=log,
            log ticks with fixed point,
            axis y line*=left,
            ymin=0, ymax=1,
            xmin=0.01, xmax=10,
            xmajorgrids,
            xminorgrids,
            xlabel=Noise level $\alpha$,
            ylabel=$\mathrm{ACC}^{\delta_m}$,
            every axis y label/.style={
    at={(ticklabel* cs:1.05)},
    anchor=south,
}]
\addplot[black, mark=o] coordinates {
(0.01, 0.86)
(0.02, 0.843)
(0.03, 0.848)
(0.04, 0.854)
(0.05, 0.856)
(0.06, 0.865)
(0.07, 0.864)
(0.08, 0.862)
(0.09, 0.864)
(0.1, 0.862)
(0.2, 0.858)
(0.3, 0.82)
(0.4, 0.792)
(0.5, 0.753)
(0.6, 0.718)
(0.7, 0.691)
(0.8, 0.642)
(0.9, 0.614)
(1, 0.58)
(2, 0.424)
(3, 0.352)
(4, 0.322)
(5, 0.304)
(6, 0.289)
(7, 0.25)
(8, 0.26)
(9, 0.252)
(10, 0.248)
    };

\label{plot_one_bottom}
\addplot[gray, mark=diamond] coordinates {
(0.010000,0.936000)
(0.020000,0.855000)
(0.030000,0.780000)
(0.040000,0.726000)
(0.050000,0.681000)
(0.060000,0.628000)
(0.070000,0.579000)
(0.080000,0.536000)
(0.090000,0.505000)
(0.100000,0.455000)
(0.200000,0.193000)
(0.300000,0.137000)
(0.400000,0.110000)
(0.500000,0.110000)
(0.600000,0.111000)
(0.700000,0.111000)
(0.800000,0.109000)
(0.900000,0.109000)
(1.000000,0.110000)
(2.000000,0.110000)
(3.000000,0.110000)
(4.000000,0.110000)
(5.000000,0.110000)
(6.000000,0.110000)
(7.000000,0.110000)
(8.000000,0.110000)
(9.000000,0.110000)
(10.000000,0.110000)
    };
\addplot[gray,smooth, mark=none, dashed] coordinates {
(0.01,0.1)
(10,0.1)
    };
\label{Random_bottom}
\end{axis}  
\begin{axis}[scale=0.78,xmode=log,
            log ticks with fixed point,
            axis y line*=right,
            ymin=3, ymax=14,
            xmin=0.01, xmax=10,
            axis x line=none,
            ylabel=\textcolor{red}{$\delta_m$},every axis y label/.style={
    at={(ticklabel* cs:1.05)},
    anchor=south,
},
            legend style={at={(-0.8,0.5)},anchor=west},
            legend style={nodes={scale=0.9, transform shape}}]
                        \addlegendimage{mark=o}
\addlegendentry{$\mathrm{ACC}_g$}

\addlegendimage{mark=diamond, gray}
\addlegendentry{$\mathrm{ACC}_f$}
\addlegendimage{mark=none, gray, dashed}
\addlegendentry{Random}
\addplot[red, mark=x] coordinates {
(0.01, 5.13843227289422)
(0.02, 5.320820114861008)
(0.03, 5.356326515140981)
(0.04, 5.428852160542549)
(0.05, 5.426275500481043)
(0.06, 5.439599361874396)
(0.07, 5.47820054953225)
(0.08, 5.512434300836492)
(0.09, 5.541754635850517)
(0.1, 5.603877971983593)
(0.2, 6.089844562158716)
(0.3, 6.7362646002776305)
(0.4, 7.259281421727414)
(0.5, 7.751918692584218)
(0.6, 8.147375623722022)
(0.7, 8.546365385139183)
(0.8, 8.95815753103678)
(0.9, 9.271474814659381)
(1, 9.534718735531053)
(2, 11.098392030963222)
(3, 11.68659998114136)
(4, 11.903514466306813)
(5, 12.061988360544959)
(6, 12.194142440331637)
(7, 12.253005523489879)
(8, 12.359447044793413)
(9, 12.400683403407282)
(10, 12.44947416569273)
    };
\addplot[orange, mark=|] coordinates {
(0.01, 11.850640473951454)
(0.02, 12.104642325673865)
(0.03, 12.16674499271645)
(0.04, 12.366398858146605)
(0.05, 12.38369122028657)
(0.06, 12.499865528191124)
(0.07, 12.152372540347033)
(0.08, 12.154481108533457)
(0.09, 12.09892442012627)
(0.1, 12.350405457656741)
(0.2, 12.501815281206083)
(0.3, 12.657277689804022)
(0.4, 12.832628358686962)
(0.5, 12.768949735499913)
(0.6, 12.650596956457626)
(0.7, 12.710893394408767)
(0.8, 12.698343540237)
(0.9, 12.641019627774488)
(1, 12.516100614840266)
(2, 12.682395709173784)
(3, 12.820398645000614)
(4, 12.84265931308978)
(5, 12.904110450486819)
(6, 12.955113610229388)
(7, 12.879292300592565)
(8, 12.98770960640464)
(9, 13.007691650775289)
(10, 13.03407446079167)
    };
\label{std_false_bottom}
\addplot[orange, mark=-] coordinates {
(0.01, 4.045747216908158)
(0.02, 4.0574036414356)
(0.03, 4.135591127651038)
(0.04, 4.242807877345603)
(0.05, 4.25586911771002)
(0.06, 4.337708110483924)
(0.07, 4.427636439866959)
(0.08, 4.449090380346723)
(0.09, 4.509607540177481)
(0.1, 4.523807446434992)
(0.2, 5.028655934997031)
(0.3, 5.436530019649886)
(0.4, 5.7955741453542)
(0.5, 6.106225906926614)
(0.6, 6.378700949862077)
(0.7, 6.684080066956402)
(0.8, 6.872508634940707)
(0.9, 7.153161625958353)
(1, 7.375787029824383)
(2, 8.946537977545097)
(3, 9.599379713582278)
(4, 9.92606041003709)
(5, 10.133971996730702)
(6, 10.321995375289074)
(7, 10.374145192181825)
(8, 10.571315138669151)
(9, 10.598928764394305)
(10, 10.676815206360459)
    };
\label{std_true_bottom}
\addlegendimage{/pgfplots/refstyle=plot_one_bottom}\addlegendentry{$\delta_m$}
\addlegendimage{/pgfplots/refstyle=std_false_bottom}\addlegendentry{$\delta_{m_{\mathrm{False}}}$}
\addlegendimage{/pgfplots/refstyle=std_true_bottom}\addlegendentry{$\delta_{m_{\mathrm{True}}}$}

  \end{axis}
\end{tikzpicture}

	\end{minipage}
	\hspace{1.4cm}
	\begin{minipage}[t]{0.3\textwidth}
		\pgfplotsset{compat=1.3,small}
\begin{tikzpicture}[baseline, scale=0.78]
\begin{axis}[scale=0.78,axis y line*=left,
            ymin=0, ymax=1,
            xmin=1, xmax=14,
            xmajorgrids,
            xminorgrids,
            title=\Large{(b)},
            xlabel=Masking level $\beta$,
            ylabel=$\mathrm{ACC}^{h}$,
            every axis y label/.style={
    at={(ticklabel* cs:1.05)},
    anchor=south,
},
            legend style={nodes={scale=0.78, transform shape}}]
\addplot[black, mark=o] coordinates {
(1, 0.871)
(2, 0.867)
(3, 0.87)
(4, 0.857)
(5, 0.862)
(6, 0.854)
(7, 0.806)
(8, 0.733)
(9, 0.587)
(10, 0.394)
(11, 0.321)
(12, 0.21)
(13, 0.166)
(14, 0.126)
    };

\addplot[gray, mark=diamond] coordinates {
(1.000000,0.766000)
(2.000000,0.761000)
(3.000000,0.736000)
(4.000000,0.708000)
(5.000000,0.617000)
(6.000000,0.499000)
(7.000000,0.289000)
(8.000000,0.153000)
(9.000000,0.138000)
(10.000000,0.131000)
(11.000000,0.128000)
(12.000000,0.111000)
(13.000000,0.110000)
(14.000000,0.110000)
};
\addplot[gray,smooth, mark=none, dashed] coordinates {
(1,0.1)
(14,0.1)
    };
\end{axis}  
\begin{axis}[scale=0.78,axis y line*=right,
            ymin=0, ymax=2.2,
            xmin=1, xmax=14,
            axis x line=none,
            ylabel=\textcolor{red}{$\delta_r$},
            every axis y label/.style={
    at={(ticklabel* cs:1.05)},
    anchor=south,
},
            legend style={at={(0.05,0.4)},anchor=west},
            legend style={nodes={scale=0.78, transform shape}}]

\addplot[red, mark=x] coordinates {
(1, 0.2610893484277511)
(2, 0.2613996948080348)
(3, 0.2646145313218816)
(4, 0.2730685581865933)
(5, 0.29306257571145555)
(6, 0.3278986770657249)
(7, 0.38954044823953676)
(8, 0.5037484564210757)
(9, 0.6950811638767723)
(10, 1.030388646760387)
(11, 1.4349993515124824)
(12, 1.7649964245931469)
(13, 1.9972613503703032)
(14, 2.072075877528459)
    };

    \addplot[orange, mark=|] coordinates {
(1, 0.29018467134154385)
(2, 0.29592358829378645)
(3, 0.2884474233625749)
(4, 0.28722087028932725)
(5, 0.3010051088372677)
(6, 0.35841567807537006)
(7, 0.4036106060784774)
(8, 0.5005323376579797)
(9, 0.7218042244456162)
(10, 1.0489539852520668)
(11, 1.4458587311834505)
(12, 1.769889377604653)
(13, 1.9969107530416443)
(14, 2.071593832039184)
    };
\addplot[orange, mark=-] coordinates {
(1, 0.25678016742214926)
(2, 0.25610364194343854)
(3, 0.2610532945801688)
(4, 0.27070708720562364)
(5, 0.2917910332852815)
(6, 0.322681484855645)
(7, 0.38615383456614416)
(8, 0.504919948521685)
(9, 0.6762794193879604)
(10, 1.0018338367960264)
(11, 1.4120288879717116)
(12, 1.746589601359386)
(13, 1.9990227851419995)
(14, 2.0754195898905716)
    };
  \end{axis}
\end{tikzpicture}\\
\begin{tikzpicture}[baseline, scale=0.78]
\begin{axis}[scale=0.78,axis y line*=left,
            ymin=0, ymax=1,
            xmin=1, xmax=14,
            xmajorgrids,
            xminorgrids,
            xlabel=Masking level $\beta$,
            ylabel=$\mathrm{ACC}^{\delta_m}$,
            every axis y label/.style={
    at={(ticklabel* cs:1.05)},
    anchor=south,
},
            legend style={nodes={scale=0.78, transform shape}},]

\addplot[black, mark=o] coordinates {
(1, 0.866)
(2, 0.861)
(3, 0.862)
(4, 0.851)
(5, 0.841)
(6, 0.829)
(7, 0.778)
(8, 0.659)
(9, 0.482)
(10, 0.299)
(11, 0.193)
(12, 0.16)
(13, 0.142)
(14, 0.115)
    };
\addplot[gray, mark=diamond] coordinates {
(1.000000,0.762000)
(2.000000,0.756000)
(3.000000,0.730000)
(4.000000,0.700000)
(5.000000,0.608000)
(6.000000,0.487000)
(7.000000,0.275000)
(8.000000,0.150000)
(9.000000,0.135000)
(10.000000,0.111000)
(11.000000,0.110000)
(12.000000,0.110000)
(13.000000,0.110000)
(14.000000,0.110000)
    };
\addplot[gray,smooth, mark=none, dashed] coordinates {
(0,0.1)
(14,0.1)
    };
\end{axis}  
\begin{axis}[scale=0.78,axis y line*=right,
            ymin=3, ymax=17.5,
            xmin=1, xmax=14,
            axis x line=none,
            ylabel=\textcolor{red}{$\delta_m$},
            every axis y label/.style={
    at={(ticklabel* cs:1.05)},
    anchor=south,
},
            legend style={at={(0.01,0.4)},anchor=west},
            legend style={nodes={scale=0.78, transform shape}},]
\addplot[red, mark=x] coordinates {
(1, 5.5846937824643526)
(2, 5.649589056348564)
(3, 5.6160404930997165)
(4, 5.729722398821481)
(5, 5.87001664678334)
(6, 6.300607839274489)
(7, 7.161627904927534)
(8, 8.73866704344245)
(9, 11.170564281530737)
(10, 14.408680743386073)
(11, 16.16476758260979)
(12, 16.31917491177355)
(13, 15.819140792413632)
(14, 15.399848278998837)
    };

\addplot[orange, mark=|] coordinates {
(1, 12.239018713872184)
(2, 12.39472170835417)
(3, 12.083315443410266)
(4, 11.906713923364743)
(5, 12.072713850419913)
(6, 12.497570977966847)
(7, 13.117857458514939)
(8, 13.715356792226135)
(9, 14.709092859062412)
(10, 16.286807391956764)
(11, 17.059778319289272)
(12, 16.850337237224778)
(13, 16.24803318647567)
(14, 15.744300623379031)
    };

\addplot[orange, mark=-] coordinates {
(1, 4.555040733031732)
(2, 4.560653587557879)
(3, 4.58067628991775)
(4, 4.648204493819194)
(5, 4.697330730756925)
(6, 5.022344031413943)
(7, 5.462035410202079)
(8, 6.163475534587764)
(9, 7.3677472625236655)
(10, 10.005447363292243)
(11, 12.422416989343763)
(12, 13.53057270315461)
(13, 13.22766421420781)
(14, 12.749062846159953)
    };
\end{axis}

\end{tikzpicture}
	\end{minipage}
	\hspace{-0.45cm}
	\begin{minipage}[t]{0.3\textwidth}
		\pgfplotsset{compat=1.3,small}
\begin{tikzpicture}[baseline, scale=0.78]
\begin{axis}[scale=0.78,axis y line*=left,
            ymin=0, ymax=1,
            xmin=0.1, xmax=10,
            xmode=log,
            xmajorgrids,
            xminorgrids,
            	title=\Large{(c)},
            xlabel=Convolution level $\gamma$,
            ylabel=$\mathrm{ACC}^h$,
            every axis y label/.style={
    at={(ticklabel* cs:1.05)},
    anchor=south,
},
            legend style={nodes={scale=0.5, transform shape}}]
\addplot[black, mark=o] coordinates {
(0.1, 0.863)
(0.2, 0.859)
(0.3, 0.869)
(0.4, 0.863)
(0.5, 0.861)
(0.6, 0.855)
(0.7, 0.871)
(0.8, 0.867)
(0.9, 0.863)
(1, 0.86)
(1.1, 0.857)
(1.2, 0.85)
(1.3, 0.833)
(1.4, 0.819)
(1.5, 0.814)
(1.6, 0.777)
(1.7, 0.758)
(1.8, 0.736)
(1.9, 0.718)
(2, 0.727)
(3, 0.596)
(4, 0.539)
(5, 0.536)
(6, 0.521)
(7, 0.52)
(8, 0.533)
(9, 0.525)
(10, 0.515)
    };
\addplot[gray, mark=diamond] coordinates {
(0.1, 0.239)
(0.2, 0.239)
(0.3, 0.241)
(0.4, 0.238)
(0.5, 0.242)
(0.6, 0.242)
(0.7, 0.247)
(0.8, 0.247)
(0.9, 0.244)
(1, 0.243)
(1.1, 0.243)
(1.2, 0.238)
(1.3, 0.232)
(1.4, 0.224)
(1.5, 0.216)
(1.6, 0.22)
(1.7, 0.214)
(1.8, 0.208)
(1.9, 0.205)
(2, 0.198)
(3, 0.184)
(4, 0.173)
(5, 0.172)
(6, 0.17)
(7, 0.167)
(8, 0.165)
(9, 0.164)
(10, 0.164)
};
\addplot[gray,smooth, mark=none, dashed] coordinates {
(0.1,0.1)
(10,0.1)
    };
\end{axis}  
\begin{axis}[scale=0.78,axis y line*=right,
            ymin=0, ymax=1.3,
            xmin=0.1, xmax=10,
            axis x line=none,
            xmode=log,
            ylabel=\textcolor{red}{$\delta_r$},
            every axis y label/.style={
    at={(ticklabel* cs:1.05)},
    anchor=south,
},
            legend style={at={(0.05,0.4)},anchor=west},
            legend style={nodes={scale=0.5, transform shape}}]
\addplot[red, mark=x] coordinates {
(0.1, 0.26710591270020256)
(0.2, 0.26667468625942115)
(0.3, 0.2696304962012446)
(0.4, 0.28699117665052426)
(0.5, 0.3146366476798152)
(0.6, 0.34767632308970997)
(0.7, 0.3776717913146803)
(0.8, 0.4085029239886046)
(0.9, 0.43642100424899827)
(1, 0.46885605141109316)
(1.1, 0.5035361486057753)
(1.2, 0.5370343920316443)
(1.3, 0.5746353363083706)
(1.4, 0.6110984851858429)
(1.5, 0.651694915664365)
(1.6, 0.6839071964557495)
(1.7, 0.7156698000770939)
(1.8, 0.7526756139899446)
(1.9, 0.7794766797451299)
(2, 0.8035521728430022)
(3, 0.9441929209938785)
(4, 0.9909509852090657)
(5, 1.0040873798767995)
(6, 1.011986893862207)
(7, 1.0212981082839823)
(8, 1.0149530340168795)
(9, 1.015880895052075)
(10, 1.0240555922705772)
    };
    \addplot[orange, mark=|] coordinates {
    (0.1, 0.259478213266207)
(0.2, 0.2664152933095628)
(0.3, 0.2637093056465795)
(0.4, 0.2882339846026867)
    (0.5, 0.31641116932335517)
(0.6, 0.36083578109651576)
(0.7, 0.38254207440109855)
(0.8, 0.43355684535508177)
(0.9, 0.4455636563095586)
(1, 0.4952264783786314)
(1.1, 0.5486093609992694)
(1.2, 0.5899777885363489)
(1.3, 0.6366460085246732)
(1.4, 0.6695481008138467)
(1.5, 0.7392972663405356)
(1.6, 0.7588556645241266)
(1.7, 0.7945970949119179)
(1.8, 0.8196581210488504)
(1.9, 0.8538582065668463)
(2, 0.8907115994175565)
(3, 1.0030425972806083)
(4, 1.0384215462668496)
(5, 1.0493566249632225)
(6, 1.047767428407217)
(7, 1.0524468143633627)
(8, 1.0460355090083393)
(9, 1.0560197259499982)
(10, 1.0564028942437063)
    };
\addplot[orange, mark=-] coordinates {
(0.1, 0.26831679893711724)
(0.2, 0.2667172641475818)
(0.3, 0.2705231037532136)
(0.4, 0.2867938826882459)
(0.5, 0.314350168575922)
(0.6, 0.3454446021411874)
(0.7, 0.3769504749907447)
(0.8, 0.40465958887702275)
(0.9, 0.4349696214769278)
(1, 0.4645631912070753)
(1.1, 0.4960151808435003)
(1.2, 0.5276914397072846)
(1.3, 0.5622034248316329)
(1.4, 0.5981810487650019)
(1.5, 0.6316776709152646)
(1.6, 0.6623968896613504)
(1.7, 0.6904713761324667)
(1.8, 0.7286492799362067)
(1.9, 0.7502627653109739)
(2, 0.7708224294388022)
(3, 0.9043015296854241)
(4, 0.9503500044156733)
(5, 0.9648990781601946)
(6, 0.9790907786087331)
(7, 0.9925454565184)
(8, 0.9877194208442497)
(9, 0.9795648099539541)
(10, 0.9935925991502519)
    };
  \end{axis}
\end{tikzpicture}
\begin{tikzpicture}[baseline, scale=0.78]
\begin{axis}[scale=0.78,axis y line*=left,
            ymin=0, ymax=1,
            xmin=0.1, xmax=10,
            	xmode=log,
            xmajorgrids,
            xminorgrids,
            xlabel=Convolution level $\gamma$,
            ylabel=$\mathrm{ACC}^{\delta_m}$,
            every axis y label/.style={
    at={(ticklabel* cs:1.05)},
    anchor=south,
},
            legend style={nodes={scale=0.5, transform shape}},]
\addplot[black, mark=o] coordinates {
(0.1, 0.844)
(0.2, 0.836)
(0.3, 0.844)
(0.4, 0.847)
(0.5, 0.837)
(0.6, 0.834)
(0.7, 0.833)
(0.8, 0.822)
(0.9, 0.83)
(1, 0.813)
(1.1, 0.81)
(1.2, 0.791)
(1.3, 0.777)
(1.4, 0.747)
(1.5, 0.755)
(1.6, 0.713)
(1.7, 0.693)
(1.8, 0.675)
(1.9, 0.659)
(2, 0.664)
(3, 0.549)
(4, 0.491)
(5, 0.493)
(6, 0.492)
(7, 0.473)
(8, 0.497)
(9, 0.481)
(10, 0.482)
    };
\addplot[gray, mark=diamond] coordinates {
(0.1, 0.226)
(0.2, 0.226)
(0.3, 0.226)
(0.4, 0.223)
(0.5, 0.232)
(0.6, 0.229)
(0.7, 0.227)
(0.8, 0.228)
(0.9, 0.228)
(1, 0.233)
(1.1, 0.234)
(1.2, 0.23)
(1.3, 0.229)
(1.4, 0.221)
(1.5, 0.219)
(1.6, 0.216)
(1.7, 0.212)
(1.8, 0.205)
(1.9, 0.203)
(2, 0.2)
(3, 0.189)
(4, 0.177)
(5, 0.173)
(6, 0.169)
(7, 0.169)
(8, 0.168)
(9, 0.166)
(10, 0.166)
    };
    \addplot[gray,smooth, mark=none, dashed] coordinates {
(0.1,0.1)
(10,0.1)
    };
\end{axis}  
\begin{axis}[scale=0.78,axis y line*=right,
            ymin=4, ymax=16,
            xmin=0.1, xmax=10,
            xmode=log,
            axis x line=none,
            ylabel=\textcolor{red}{$\delta_m$},
            every axis y label/.style={
    at={(ticklabel* cs:1.05)},
    anchor=south,
},
            legend style={at={(0.01,0.4)},anchor=west},
            legend style={nodes={scale=0.5, transform shape}},]

\addplot[red, mark=x] coordinates {
(0.1, 5.8429204273655)
(0.2, 5.8836218521007355)
(0.3, 5.835267279903696)
(0.4, 5.930187649945964)
(0.5, 6.145132239322621)
(0.6, 6.434529938807662)
(0.7, 6.576325416274937)
(0.8, 6.834243385402113)
(0.9, 7.071938663676547)
(1, 7.326936473749593)
(1.1, 7.623418835540388)
(1.2, 7.923812753136907)
(1.3, 8.273323133552331)
(1.4, 8.657065220318644)
(1.5, 8.982956560968086)
(1.6, 9.312810940396835)
(1.7, 9.645472783476107)
(1.8, 10.015903390541574)
(1.9, 10.345769856060189)
(2, 10.561640454760637)
(3, 12.272865718451872)
(4, 12.77211968858815)
(5, 12.93706738241974)
(6, 13.066429544327763)
(7, 13.142680641348216)
(8, 13.047047141778133)
(9, 13.038126373677112)
(10, 13.142448324539684)
    };
\addplot[orange, mark=|] coordinates {
(0.1, 12.193429823533885)
(0.2, 12.239572087758301)
(0.3, 12.098846625106507)
(0.4, 12.198058872140507)
(0.5, 12.323758928844676)
(0.6, 12.668355424415081)
(0.7, 12.68815978012218)
(0.8, 12.76933777841388)
(0.9, 13.504485521035491)
(1, 13.62630659518774)
(1.1, 13.806100588578774)
(1.2, 13.886419271531793)
(1.3, 13.919572317769545)
(1.4, 13.809832366376508)
(1.5, 14.147121183805226)
(1.6, 14.035961914294463)
(1.7, 13.978248623872817)
(1.8, 13.927748517474056)
(1.9, 14.302253157655578)
(2, 14.49610086484902)
(3, 15.066277685384422)
(4, 15.112271199378473)
(5, 15.355264009450227)
(6, 15.392109267177572)
(7, 15.267790736340498)
(8, 15.337616140484004)
(9, 15.368556612519496)
(10, 15.348711895766183)
    };
\addplot[orange, mark=-] coordinates {
(0.1, 4.669129591106889)
(0.2, 4.636760801086573)
(0.3, 4.677544083397015)
(0.4, 4.797974784543643)
(0.5, 4.941887137301002)
(0.6, 5.193744530401389)
(0.7, 5.351023689069067)
(0.8, 5.549028297864285)
(0.9, 5.7544290663861615)
(1, 5.878003862791496)
(1.1, 6.173160152728915)
(1.2, 6.348357933485161)
(1.3, 6.652842350952023)
(1.4, 6.911884379685927)
(1.5, 7.307168040974578)
(1.6, 7.411626747537623)
(1.7, 7.7260468339785735)
(1.8, 8.13242240350001)
(1.9, 8.298484869953924)
(2, 8.570708680980973)
(3, 9.978095596254095)
(4, 10.34617850937782)
(5, 10.450199857258568)
(6, 10.665117960572267)
(7, 10.774957554538634)
(8, 10.728825398621083)
(9, 10.52358730099687)
(10, 10.771401582018258)
    };
    \end{axis}
\end{tikzpicture}
	\end{minipage}

%% file: images/figure4.tex
	    \pgfplotsset{
            compat=1.3,
            layers/my layer set/.define layer set={
            background,
            main,
            foreground, 
        }{},
 	        set layers=my layer set,
    }
    
\begin{tikzpicture}[scale=0.8]
    \begin{axis}[ticklabel style = {font=\tiny}, label style={font=\tiny},xlabel=PC-1, ylabel=PC-2,legend style={nodes={scale=0.7, transform shape}},legend style={at={(1.3,0.7)},anchor=west}, scale=0.8]
\addplot[on layer=main,
scatter,
only marks,
point meta=explicit symbolic,
scatter/classes={
0={mark=square,scale=0.4,Red},
1={mark=x,scale=0.4,Blue},
2={mark=+,scale=0.4,Green},
3={mark=pentagon,scale=0.4,Brown},
4={mark=o,scale=0.4,Orange},
5={mark=Mercedes star,scale=0.4,Gray},
6={mark=star,scale=0.4,Purple},
7={mark=diamond,scale=0.4,Cyan},
8={mark=oplus,scale=0.4,VioletRed},
9={mark=triangle,scale=0.4,Black}},
]
table[col sep=comma, meta=label] {images/figure3/data.csv};
\addplot[black, mark=o,
only marks]coordinates {
(-3,-2)}node[pin=150:{\tiny{$\overline{\mathcal{H}}$}}]{};
\addplot[black,-latex,on layer=foreground,line width=0.7pt]coordinates {
(-3,-2)
(1.4368732, 0.29129496)
}node[pos=0.8,pin={[pin edge={solid,-}]75:{\tiny{$\delta_{m_0}$}}}]{};
\addplot[black,-latex,on layer=foreground,line width=0.7pt]coordinates {
(-3,-2)
(-2.7275455, -2.7784886)
}node[pos=0.1,pin={[pin edge={solid,-}]180:{\tiny{$\delta_{m_1}$}}}]{};
\addplot[black,-latex,on layer=foreground,line width=0.7pt]coordinates {
(-3,-2)
(-4.1129346, 5.947219)
}node[pos=0.6,pin={[pin edge={solid,-}]160:{\tiny{$\delta_{m_2}$}}}]{};
\addplot[black,-latex,on layer=foreground,line width=0.7pt]coordinates {
(-3,-2)
(5.8840013, 2.9335325)
}node[pos=0.8,pin={[pin edge={solid,-}]85:{\tiny{$\delta_{m_3}$}}}]{};
\addplot[black,-latex,on layer=foreground,line width=0.7pt]coordinates {
(-3,-2)
(-1.0261184, -0.68840045)
}node[pos=0.5,pin={[pin edge={solid,-}]90:{\tiny{$\delta_{m_4}$}}}]{};
\addplot[black,-latex,on layer=foreground,line width=0.7pt]coordinates {
(-3,-2)
(1.5054904, -1.1637026)
}node[pos=0.7,pin={[pin edge={solid,-}]30:{\tiny{$\delta_{m_5}$}}}]{};
\addplot[black,-latex,on layer=foreground,line width=0.7pt]coordinates {
(-3,-2)
(-1.4515756, -1.1309247)
}node[pos=0.9,pin={[pin edge={solid,-}]330:{\tiny{$\delta_{m_6}$}}}]{};
\addplot[black,-latex,on layer=foreground,line width=0.7pt]coordinates {
(-3,-2)
(-0.6897279, 1.1449578)
}node[pos=0.6,pin={[pin edge={solid,-}]80:{\tiny{$\delta_{m_7}$}}}]{};
\addplot[black,-latex,on layer=foreground,line width=0.7pt]coordinates {
(-3,-2)
(2.0851743, -1.3654587)
}node[pos=0.8,pin={[pin edge={solid,-}]330:{\tiny{$\delta_{m_8}$}}}]{};
\addplot[black,-latex,on layer=foreground,line width=0.7pt]coordinates {
(-3,-2)
(-0.50932276, -3.4935505)
}node[pos=0.5,pin={[pin edge={solid,-}]200:{\tiny{$\delta_{m_9}$}}}]{};
\addlegendentry{$0$}
\addlegendentry{$1$}
\addlegendentry{$2$}
\addlegendentry{$3$}
\addlegendentry{$4$}
\addlegendentry{$5$}
\addlegendentry{$6$}
\addlegendentry{$7$}
\addlegendentry{$8$}
\addlegendentry{$9$}
\end{axis}
\end{tikzpicture}

%% file: images/figure5.tex
	\begin{minipage}{0.6\textwidth} %
		\subfloat{\pgfplotsset{compat=1.3,small}
\begin{tikzpicture}[scale=0.8]
\begin{axis}[scale=0.8,
			axis y line*=left,
            ymin=0, ymax=1,
            xmin=0, xmax=1,
            xmajorgrids,
            xminorgrids,
            xlabel=False Positive Rate,
            ylabel=True Positive Rate,
            	title=U-ROC-Curve $\alpha \equal 0.1$, 
            width=6.5cm, height=5.5cm,
            legend style={at={(1.4,0.5)},anchor=west}, 
            legend style={nodes={scale=0.7, transform shape}}, legend cell align={left}]
\addplot[black, mark=o] coordinates {
(0.0,0.0)
(0.0,0.002320185614849188)
(0.0,0.002320185614849188)
(0.0,0.0069605568445475635)
(0.0,0.019721577726218097)
(0.0,0.04524361948955916)
(0.0,0.08352668213457076)
(0.0,0.12296983758700696)
(0.0,0.1716937354988399)
(0.0,0.22621809744779584)
(0.0,0.28306264501160094)
(0.0,0.351508120649652)
(0.0,0.40835266821345706)
(0.0,0.4779582366589327)
(0.007246376811594203,0.5382830626450116)
(0.021739130434782608,0.5823665893271461)
(0.036231884057971016,0.62877030162413)
(0.043478260869565216,0.6647331786542924)
(0.050724637681159424,0.7018561484918794)
(0.050724637681159424,0.7354988399071926)
(0.07971014492753623,0.7645011600928074)
(0.10144927536231885,0.7865429234338747)
(0.12318840579710146,0.8051044083526682)
(0.14492753623188406,0.8225058004640371)
(0.16666666666666666,0.8387470997679815)
(0.1956521739130435,0.851508120649652)
(0.30434782608695654,0.8747099767981439)
(0.34782608695652173,0.8874709976798144)
(0.41304347826086957,0.9002320185614849)
(0.43478260869565216,0.9095127610208816)
(0.463768115942029,0.9245939675174014)
(0.5,0.9385150812064965)
(0.5289855072463768,0.9477958236658933)
(0.5869565217391305,0.9535962877030162)
(0.6231884057971014,0.9582366589327146)
(0.6594202898550725,0.9605568445475638)
(0.6884057971014492,0.9651972157772621)
(0.7318840579710145,0.9709976798143851)
(0.7463768115942029,0.974477958236659)
(0.7536231884057971,0.9756380510440835)
(0.7753623188405797,0.9802784222737819)
(0.782608695652174,0.9837587006960556)
(0.8043478260869565,0.9849187935034803)
(0.8188405797101449,0.9849187935034803)
(0.8260869565217391,0.9895591647331786)
(0.8333333333333334,0.9895591647331786)
(0.855072463768116,0.9895591647331786)
(0.855072463768116,0.9907192575406032)
(0.855072463768116,0.9930394431554525)
(0.8913043478260869,0.994199535962877)
(0.8985507246376812,0.994199535962877)
(0.9130434782608695,0.9953596287703016)
(0.927536231884058,0.9965197215777262)
(1.0,1.0)
    };

\addplot[cyan, mark=x] coordinates {
(0.0,0.0)
(0.0,0.0011587485515643105)
(0.0,0.0011574074074074073)
(0.0,0.002325581395348837)
(0.0,0.0)
(0.0,0.0011614401858304297)
(0.0,0.0)
(0.0,0.005813953488372093)
(0.0,0.003464203233256351)
(0.0,0.0069605568445475635)
(0.0,0.0069605568445475635)
(0.0,0.017341040462427744)
(0.0,0.011614401858304297)
(0.0,0.016203703703703703)
(0.0,0.01853997682502897)
(0.0,0.031286210892236384)
(0.0,0.025433526011560695)
(0.007042253521126761,0.03496503496503497)
(0.007352941176470588,0.04050925925925926)
(0.0,0.03841676367869616)
(0.0,0.0429732868757259)
(0.0,0.04519119351100811)
(0.0,0.06936416184971098)
(0.0,0.06511627906976744)
(0.007407407407407408,0.09364161849710982)
(0.007518796992481203,0.0922722029988466)
(0.0,0.10788863109048724)
(0.0072992700729927005,0.11471610660486674)
(0.014598540145985401,0.13557358053302435)
(0.0,0.13921113689095127)
(0.014814814814814815,0.1699421965317919)
(0.030303030303030304,0.16589861751152074)
(0.0,0.19674039580908032)
(0.037037037037037035,0.20346820809248556)
(0.014285714285714285,0.24302325581395348)
(0.021739130434782608,0.23781902552204176)
(0.022058823529411766,0.2534722222222222)
(0.05,0.28255813953488373)
(0.051094890510948905,0.30243337195828507)
(0.04964539007092199,0.3410942956926659)
(0.07246376811594203,0.37935034802784223)
(0.04895104895104895,0.40606767794632437)
(0.11188811188811189,0.4200700116686114)
(0.08695652173913043,0.46867749419953597)
(0.10218978102189781,0.5052143684820394)
(0.13286713286713286,0.5262543757292882)
(0.20588235294117646,0.6018518518518519)
(0.2302158273381295,0.6445993031358885)
(0.28888888888888886,0.6763005780346821)
(0.4460431654676259,0.7630662020905923)
(0.5912408759124088,0.8424101969872537)
(1.0,1.0)
    };

\addplot[orange, mark=|] coordinates {
(0.0,0.0)
(0.0,0.006006006006006006)
(0.0,0.021021021021021023)
(0.0,0.06456456456456457)
(0.0,0.11861861861861862)
(0.0,0.16366366366366367)
(0.0,0.21021021021021022)
(0.0029940119760479044,0.25375375375375375)
(0.008982035928143712,0.2927927927927928)
(0.011976047904191617,0.32732732732732733)
(0.017964071856287425,0.35435435435435436)
(0.03592814371257485,0.3963963963963964)
(0.038922155688622756,0.42492492492492495)
(0.059880239520958084,0.44894894894894893)
(0.059880239520958084,0.466966966966967)
(0.08383233532934131,0.487987987987988)
(0.09580838323353294,0.4984984984984985)
(0.10778443113772455,0.5075075075075075)
(0.10778443113772455,0.5180180180180181)
(0.12574850299401197,0.5315315315315315)
(0.1407185628742515,0.5480480480480481)
(0.1467065868263473,0.5570570570570571)
(0.15568862275449102,0.5645645645645646)
(0.16766467065868262,0.5675675675675675)
(0.17365269461077845,0.5735735735735735)
(0.18562874251497005,0.5825825825825826)
(0.18562874251497005,0.5900900900900901)
(0.20658682634730538,0.5960960960960962)
(0.2155688622754491,0.6051051051051051)
(0.218562874251497,0.6141141141141141)
(0.24251497005988024,0.6276276276276276)
(0.25149700598802394,0.6351351351351351)
(0.26047904191616766,0.6411411411411412)
(0.2694610778443114,0.6471471471471472)
(0.281437125748503,0.6561561561561562)
(0.2964071856287425,0.6681681681681682)
(0.30538922155688625,0.6726726726726727)
(0.3083832335329341,0.6771771771771772)
(0.32335329341317365,0.6786786786786787)
(0.3323353293413174,0.6876876876876877)
(0.33532934131736525,0.6936936936936937)
(0.3383233532934132,0.6951951951951952)
(0.344311377245509,0.6996996996996997)
(0.3502994011976048,0.7042042042042042)
(0.36227544910179643,0.7102102102102102)
(0.38023952095808383,0.7177177177177178)
(0.38622754491017963,0.7222222222222222)
(0.38622754491017963,0.7267267267267268)
(0.39520958083832336,0.7267267267267268)
(0.4041916167664671,0.7267267267267268)
(0.4101796407185629,0.7312312312312312)
(0.4101796407185629,0.7327327327327328)
(0.4161676646706587,0.7357357357357357)
(0.4251497005988024,0.7387387387387387)
(0.4341317365269461,0.7447447447447447)
(0.4491017964071856,0.7492492492492493)
(0.45209580838323354,0.7537537537537538)
(0.45209580838323354,0.7567567567567568)
(0.46107784431137727,0.7582582582582582)
(0.46107784431137727,0.7597597597597597)
(0.47305389221556887,0.7612612612612613)
(0.4820359281437126,0.7657657657657657)
(0.4820359281437126,0.7672672672672672)
(0.48502994011976047,0.7702702702702703)
(0.48502994011976047,0.7747747747747747)
(0.5,0.7792792792792793)
(0.5089820359281437,0.7792792792792793)
(0.5299401197604791,0.7822822822822822)
(0.5299401197604791,0.7852852852852853)
(0.5359281437125748,0.7897897897897898)
(0.5359281437125748,0.7912912912912913)
(0.5389221556886228,0.7942942942942943)
(0.5419161676646707,0.7957957957957958)
(0.5508982035928144,0.7972972972972973)
(0.5538922155688623,0.7987987987987988)
(0.562874251497006,0.8018018018018018)
(0.5688622754491018,0.8048048048048048)
(0.5688622754491018,0.8063063063063063)
(0.5778443113772455,0.8078078078078078)
(0.5808383233532934,0.8093093093093093)
(0.592814371257485,0.8108108108108109)
(0.5988023952095808,0.8123123123123123)
(0.6017964071856288,0.8138138138138138)
(0.6047904191616766,0.8153153153153153)
(0.6137724550898204,0.8168168168168168)
(0.6197604790419161,0.8183183183183184)
(0.7604790419161677,0.8843843843843844)
(0.7634730538922155,0.8858858858858859)
(1.0, 1.0)
    };
\addplot[gray,smooth, mark=none, dashed] coordinates {
(0,0)
(1,1)
    };
\end{axis}
\end{tikzpicture}}
		\subfloat{\pgfplotsset{compat=1.3,small}
\begin{tikzpicture}[scale=0.8]
\begin{axis}[scale=0.8,
			axis y line*=left,
            ymin=0, ymax=1,
            xmin=0, xmax=1,
            xmajorgrids,
            xminorgrids,
            xlabel=False Positive Rate,
            ylabel=True Positive Rate,
            	title=U-ROC-Curve $\alpha \equal 0.5$, 
            width=6.5cm, height=5.5cm,
           ]
\addplot[black, mark=o] coordinates {
(0.0,0.0)
(0.0,0.0013315579227696406)
(0.0,0.0013315579227696406)
(0.0,0.002663115845539281)
(0.0,0.010652463382157125)
(0.0,0.022636484687083888)
(0.0,0.0426098535286285)
(0.0,0.06790945406125166)
(0.0,0.085219707057257)
(0.0,0.10918774966711052)
(0.0,0.14513981358189082)
(0.0,0.1810918774966711)
(0.0,0.21438082556591212)
(0.0,0.27030625832223704)
(0.0,0.30492676431424764)
(0.004016064257028112,0.35286284953395475)
(0.012048192771084338,0.39280958721704395)
(0.01606425702811245,0.45006657789613846)
(0.01606425702811245,0.4966711051930759)
(0.020080321285140562,0.5352862849533955)
(0.024096385542168676,0.5778961384820239)
(0.04417670682730924,0.625832223701731)
(0.04819277108433735,0.6524633821571239)
(0.0642570281124498,0.681757656458056)
(0.10843373493975904,0.7070572569906791)
(0.1285140562248996,0.7350199733688415)
(0.1606425702811245,0.7509986684420772)
(0.18072289156626506,0.7723035952063915)
(0.21686746987951808,0.7909454061251664)
(0.23694779116465864,0.8095872170439414)
(0.2570281124497992,0.8202396804260985)
(0.30120481927710846,0.8308921438082557)
(0.3253012048192771,0.8402130492676432)
(0.3493975903614458,0.8588548601864181)
(0.37751004016064255,0.8681757656458056)
(0.40160642570281124,0.8841544607190412)
(0.40562248995983935,0.8961384820239681)
(0.42570281124497994,0.9174434087882823)
(0.4497991967871486,0.9254327563249002)
(0.46987951807228917,0.929427430093209)
(0.4979919678714859,0.9374167776298269)
(0.5381526104417671,0.9400798934753661)
(0.5582329317269076,0.948069241011984)
(0.5863453815261044,0.9507323568575233)
(0.6144578313253012,0.9600532623169108)
(0.6305220883534136,0.9680426098535286)
(0.678714859437751,0.9707057256990679)
(0.7068273092369478,0.9720372836218375)
(0.7469879518072289,0.9773635153129161)
(0.7710843373493976,0.9800266311584553)
 (1.0,1.0)
    };

\addplot[cyan, mark=x] coordinates {
(0.0,0.0)
(0.0,0.0013054830287206266)
(0.0,0.00392156862745098)
(0.0,0.00130718954248366)
(0.0,0.003931847968545216)
(0.0,0.011795543905635648)
(0.0,0.00904392764857881)
(0.0,0.0078125)
(0.0,0.005201560468140442)
(0.0,0.016861219195849545)
(0.0,0.01568627450980392)
(0.0,0.02490170380078637)
(0.0,0.02860858257477243)
(0.004201680672268907,0.028871391076115485)
(0.0,0.025974025974025976)
(0.008658008658008658,0.04681404421326398)
(0.004201680672268907,0.03674540682414698)
(0.004310344827586207,0.036458333333333336)
(0.004201680672268907,0.05249343832020997)
(0.004273504273504274,0.06005221932114883)
(0.008733624454148471,0.0609597924773022)
(0.02092050209205021,0.06176084099868594)
(0.017316017316017316,0.0858257477243173)
(0.022321428571428572,0.09149484536082474)
(0.0205761316872428,0.09379128137384413)
(0.030434782608695653,0.10649350649350649)
(0.02643171806167401,0.11513583441138421)
(0.02575107296137339,0.1303780964797914)
(0.056768558951965066,0.14785992217898833)
(0.05333333333333334,0.1393548387096774)
(0.06896551724137931,0.18359375)
(0.07627118644067797,0.20287958115183247)
(0.13675213675213677,0.21148825065274152)
(0.14285714285714285,0.2690288713910761)
(0.15677966101694915,0.2696335078534031)
(0.19913419913419914,0.31209362808842656)
(0.2782608695652174,0.36753246753246754)
(0.35294117647058826,0.452755905511811)
(0.502262443438914,0.5430038510911425)
(1.0,1.0)
    };

\addplot[orange, mark=|] coordinates {
(0.0,0.008)
(0.0,0.048)
(0.002285714285714286,0.152)
(0.012571428571428572,0.32)
(0.038857142857142854,0.44)
(0.05714285714285714,0.544)
(0.09828571428571428,0.584)
(0.16228571428571428,0.608)
(0.21942857142857142,0.64)
(0.28114285714285714,0.664)
(0.3302857142857143,0.68)
(0.37942857142857145,0.688)
(0.4308571428571429,0.696)
(0.46514285714285714,0.704)
(0.528,0.704)
(0.5542857142857143,0.728)
(0.5725714285714286,0.736)
(0.6114285714285714,0.752)
(0.6422857142857142,0.76)
(0.6525714285714286,0.768)
(0.6628571428571428,0.776)
(0.7314285714285714,0.784)
(0.784,0.792)
(0.7874285714285715,0.8)
(0.7965714285714286,0.808)
(0.8228571428571428,0.816)
(0.8285714285714286,0.832)
(0.8422857142857143,0.84)
(0.8445714285714285,0.848)
(0.8571428571428571,0.856)
(0.864,0.864)
(0.8662857142857143,0.872)
(0.8708571428571429,0.88)
(0.9074285714285715,0.888)
(0.9302857142857143,0.928)
(1.0, 1.0)
    };
\addplot[gray,smooth, mark=none, dashed] coordinates {
(0,0)
(1,1)
    };
\end{axis}
\end{tikzpicture}
}\\
		\subfloat{\begin{tikzpicture}[scale=0.8]
\begin{axis}[scale=0.8,axis y line*=left,
            ymin=0, ymax=1,
            xmin=0, xmax=1,
            xmajorgrids,
            xminorgrids,
            xlabel=False Positive Rate,
            ylabel=True Positive Rate,
            	title=U-ROC-Curve $\alpha\equal 1.0$, 
            width=6.5cm, height=5.5cm,
             legend style={at={(1.6,0.5)},anchor=west}, 
            legend style={nodes={scale=0.9, transform shape}}
              ]
\addplot[black, mark=o] coordinates {
(0.0,0.0)
(0.0,0.0017152658662092624)
(0.0,0.008576329331046312)
(0.0,0.010291595197255575)
(0.0,0.015437392795883362)
(0.0,0.018867924528301886)
(0.0,0.032590051457975985)
(0.0,0.0411663807890223)
(0.0,0.058319039451114926)
(0.0,0.0823327615780446)
(0.0,0.09433962264150944)
(0.0,0.1269296740994854)
(0.002398081534772182,0.16123499142367068)
(0.004796163069544364,0.19039451114922812)
(0.007194244604316547,0.21612349914236706)
(0.009592326139088728,0.2469982847341338)
(0.011990407673860911,0.2847341337907376)
(0.016786570743405275,0.32246998284734135)
(0.016786570743405275,0.3567753001715266)
(0.023980815347721823,0.411663807890223)
(0.0407673860911271,0.44768439108061747)
(0.05755395683453238,0.47855917667238423)
(0.08393285371702638,0.5077186963979416)
(0.09112709832134293,0.5540308747855918)
(0.11031175059952038,0.5780445969125214)
(0.1223021582733813,0.6174957118353345)
(0.14388489208633093,0.6380789022298456)
(0.18225419664268586,0.6620926243567753)
(0.2014388489208633,0.6809605488850772)
(0.23261390887290168,0.7186963979416809)
(0.2517985611510791,0.7427101200686106)
(0.2805755395683453,0.7684391080617495)
(0.31894484412470026,0.7873070325900514)
(0.35731414868105515,0.8113207547169812)
(0.39568345323741005,0.8370497427101201)
(0.41007194244604317,0.8576329331046312)
(0.45083932853717024,0.869639794168096)
(0.4892086330935252,0.8902229845626072)
(0.5323741007194245,0.9108061749571184)
(0.565947242206235,0.9262435677530018)
(0.6019184652278178,0.9331046312178388)
(0.6258992805755396,0.9382504288164666)
(0.6666666666666666,0.9536878216123499)
(0.7074340527577938,0.9605488850771869)
(0.7362110311750599,0.9639794168096055)
(0.7889688249400479,0.9691252144082333)
(1.0, 1.0)
    };
    \addlegendentry{Our method};
\addplot[cyan, mark=x] coordinates {
(0.0,0.0)
(0.0,0.0015290519877675841)
(0.0,0.0)
(0.0,0.006144393241167435)
(0.0028735632183908046,0.0015337423312883436)
(0.0,0.0)
(0.0,0.007668711656441718)
(0.0,0.009433962264150943)
(0.0,0.0046875)
(0.0,0.007680491551459293)
(0.002824858757062147,0.01238390092879257)
(0.0028089887640449437,0.010869565217391304)
(0.0028011204481792717,0.013996889580093312)
(0.008771929824561403,0.0182370820668693)
(0.005681818181818182,0.023148148148148147)
(0.0,0.021739130434782608)
(0.003003003003003003,0.026986506746626688)
(0.008849557522123894,0.02723146747352496)
(0.014005602240896359,0.041990668740279936)
(0.008450704225352112,0.037209302325581395)
(0.008620689655172414,0.04754601226993865)
(0.011363636363636364,0.06018518518518518)
(0.028011204481792718,0.05598755832037325)
(0.023121387283236993,0.07033639143730887)
(0.0446927374301676,0.0763239875389408)
(0.04664723032069971,0.0882800608828006)
(0.05014749262536873,0.09682299546142209)
(0.06936416184971098,0.10244648318042814)
(0.07262569832402235,0.14797507788161993)
(0.09798270893371758,0.14088820826952528)
(0.11830985915492957,0.16744186046511628)
(0.12605042016806722,0.16951788491446346)
(0.17796610169491525,0.22445820433436534)
(0.1956521739130435,0.2689873417721519)
(0.3746312684365782,0.3101361573373676)
(0.5116279069767442,0.4054878048780488)
(1.0,1.0)
};
\addlegendentry{MC dropout};

\addplot[orange, mark=|] coordinates {
(0.0,0.0)
(0.0,0.0449438202247191)
(0.006586169045005488,0.23595505617977527)
(0.043907793633369926,0.5056179775280899)
(0.1350164654226125,0.7528089887640449)
(0.24039517014270034,0.8426966292134831)
(0.3358946212952799,0.9213483146067416)
(0.442371020856202,0.9662921348314607)
(0.4983534577387486,0.9662921348314607)
(0.5631174533479693,0.9775280898876404)
(0.6212952799121844,0.9887640449438202)
(0.6761800219538968,1.0)
(0.7135016465422612,1.0)
(0.7530186608122942,1.0)
(0.7793633369923162,1.0)
(0.8079034028540066,1.0)
(0.8320526893523601,1.0)
(0.845225027442371,1.0)
(0.8638858397365532,1.0)
(0.8770581778265643,1.0)
(0.889132821075741,1.0)
(0.8968166849615807,1.0)
(0.9034028540065862,1.0)
(0.9088913282107574,1.0)
(0.9154774972557629,1.0)
(0.9176728869374314,1.0)
(0.9242590559824369,1.0)
(0.9308452250274424,1.0)
(0.9319429198682766,1.0)
(0.9330406147091108,1.0)
(0.9385290889132821,1.0)
(0.9418221734357849,1.0)
(0.9429198682766191,1.0)
(0.9451152579582875,1.0)
(0.9451152579582875,1.0)
(0.9462129527991219,1.0)
(0.9462129527991219,1.0)
(0.9484083424807903,1.0)
(0.9506037321624589,1.0)
(0.9517014270032931,1.0)
(0.9560922063666301,1.0)
(0.9560922063666301,1.0)
(0.9582875960482986,1.0)
(0.960482985729967,1.0)
(0.9626783754116356,1.0)
(0.964873765093304,1.0)
(0.9659714599341384,1.0)
(0.9681668496158068,1.0)
(0.969264544456641,1.0)
(0.969264544456641,1.0)
(0.991218441273326,1.0)
(0.991218441273326,1.0)
(0.991218441273326,1.0)
(0.991218441273326,1.0)
    };
    \addlegendentry{EDL};    

\addplot[gray,smooth, mark=none, dashed] coordinates {
(0,0)
(1,1)
    };
\addlegendentry{Random}
\end{axis}
\end{tikzpicture}}
	\end{minipage} %
	\begin{minipage}{0.4\textwidth} %
	\hspace{0.5cm}
		\scalebox{0.7}{
			\begin{tabular}{llll}
				\toprule
				\multicolumn{1}{c}{\bf $\alpha$}&\multicolumn{1}{c}{\bf METHOD}  &\multicolumn{1}{c}{\bf ACC}&\multicolumn{1}{c}{\bf AUC}
				\\ \midrule
				\multirow{3}*{$0.1$} & Our Method & 0.862 & \bf 0.904\\
				& MC dropout     & \bf 0.915& 0.745
				\\
				& EDL         & 0.666    &0.746
				\\ \midrule
				\multirow{3}*{$0.5$} & Our Method & 0.753 & \bf 0.881
				\\
				& MC dropout    & \bf 0.755   &0.555
				\\          
				& EDL        & 0.125 &0.717
				\\\midrule
				\multirow{3}*{$1.0$} & Our Method  & \bf 0.580 &0.828
				\\
				& MC dropout      &0.570   & 0.472
				\\
				& EDL        & 0.089 & \bf 0.885
				\\
				\bottomrule
		\end{tabular}}
	\end{minipage}